%% file: FinalVersion_arxiv.tex
\documentclass[journal,10pt]{IEEEtran}
\usepackage[utf8]{inputenc}

\usepackage[nolist]{acronym}
\usepackage{amsmath}
\usepackage{amssymb}
\usepackage{cases}
\usepackage{mathtools}
\usepackage{gensymb} 
\usepackage{siunitx}
\usepackage{pifont}
\usepackage[ruled,vlined]{algorithm2e}
\DeclareMathOperator*{\argmax}{arg\,max}
\DeclareMathOperator*{\argmin}{arg\,min}
\DeclareMathOperator{\E}{\mathbb{E}}
\usepackage{balance}
\usepackage{blindtext, graphicx}   
\usepackage{booktabs,tabularx}
\newcolumntype{L}{>{\centering\arraybackslash}m{3cm}}
\usepackage{hyperref} 
\SetKwInput{KwInput}{Input}                
\SetKwInput{KwOutput}{Output}   
\SetAlFnt{\footnotesize} 
\LinesNumbered  
\SetAlCapSty{xAlCapSty}

\usepackage[font=small,skip=0pt]{caption}
\usepackage{enumerate}
\usepackage[list=true]{subcaption}
\usepackage[table,xcdraw]{xcolor}
\newcommand{\figref}[1]{\figurename~\ref{#1}}

\newcommand\norm[1]{\left\lVert#1\right\rVert}
\newtheorem{definition}{Definition}[section]

\begin{document}

\bstctlcite{IEEEexample:BSTcontrol}

\title{A Transfer Learning Approach for UAV Path Design with Connectivity Outage Constraint}
\author{\IEEEauthorblockN{Gianluca Fontanesi,~\IEEEmembership{Graduate Student Member,~IEEE}, Anding Zhu,~\IEEEmembership{Fellow,~IEEE},\\ Mahnaz Arvaneh,~\IEEEmembership{Member,~IEEE}, and Hamed Ahmadi,~\IEEEmembership{Senior Member,~IEEE}} 

\thanks{Manuscript received XXX, XX, 2021; revised XXX, XX, 2021.}
\thanks{This work was supported by the Irish Research Council under Grant GOIPG/2017/1741 and in part by the Science Foundation Ireland under Grant Number 17/NSFC/4850. (Corresponding author: Gianluca Fontanesi)}
\thanks{G. Fontanesi and A. Zhu are with the School of Electrical and Electronic Engineering, University College Dublin, Dublin 4, D04 V1W8, Ireland (e-mail: gianluca.fontanesi@ucdconnect.ie; anding.zhu@ucd.ie).}
\thanks{H. Ahmadi is with the Department of Electronic Engineering, University of York, Heslington, York YO10 5DD, United Kingdom, and with the School of Electrical and Electronic Engineering, University College Dublin, Dublin 4, D04 V1W8, Ireland (e-mail: hamed.ahmadi@ucd.ie).}
\thanks{M. Arvaneh is with the Automatic Control and Systems Engineering, University of Sheffield, UK(e-mail: m.arvaneh@sheffield.ac.uk)}

        }
\markboth{IEEE Internet of Things Journal,~Vol.~XX, No.~XX, XXX~2021}
{}

\maketitle

\binoppenalty=10000
\relpenalty=10000
\begin{abstract}
The connectivity-aware path design is crucial in the effective deployment of autonomous Unmanned Aerial Vehicles (UAVs). Recently, Reinforcement Learning (RL) algorithms have become the popular approach to solving this type of complex problem, but RL algorithms suffer slow convergence. In this paper, we propose a Transfer Learning (TL) approach, where we use a teacher policy previously trained in an old domain to boost the path learning of the agent in the new domain. As the exploration processes and the training continue, the agent refines the path design in the new domain based on the subsequent interactions with the environment.
We evaluate our approach considering an old domain at sub-6 GHz and a new domain at millimeter Wave (mmWave).
The teacher path policy, previously trained at sub-6 GHz path, is the solution to a connectivity-aware path problem that we formulate as a constrained Markov Decision Process (CMDP). We employ a Lyapunov-based model-free Deep Q-Network (DQN) to solve the path design at sub-6 GHz that guarantees connectivity constraint satisfaction.
We empirically demonstrate the effectiveness of our approach for different urban environment scenarios.
The results demonstrate that our proposed approach is capable of reducing the training time considerably at mmWave. 
\end{abstract}

\begin{IEEEkeywords}
Cellular networks, deep reinforcement learning, path design, transfer learning,  Unmanned Aerial Vehicle (UAV). 
\end{IEEEkeywords}
\IEEEpeerreviewmaketitle

\input{acronyms.tex}

\section{Introduction}\label{Introduction}
\acp{UAV} are expected to be a promising solution in diverse applications, such as fast delivery, surveillance and disaster management thanks to their easy deployment and high mobility \cite{zeng2016wireless}.
From the standpoint of wireless communications, on one hand, ground \acp{BS} can be leveraged to support \acp{UAV} as flying \acp{UE} for remote operations and high data rate transmissions \cite{zeng2018cellular}.
On the other hand, cellular networks can be used to provide a \ac{BH}, or \ac{FH} link to \acp{UAV} when deployed as wireless \acp{BS} or \acp{RRH}. \ac{UAV}-\acp{BS}/\acp{RRH} offer rapid deployment of on-demand communication in hotspots and provide emergency service operations \cite{rahimi2021efficient,fontanesi2019over}.

A high quality and reliable ground to air link along the entire path \cite{fontanesi2020outage} represents a crucial challenge for the effective deployment of \acp{UAV} in the above scenarios. 
An efficient \ac{UAV} path design shall thus optimize the \ac{UAV} path to minimize the travelling time and comply with the quality-of-connectivity constraint on the ground to air link.
However, designing a connectivity-aware path is particularly challenging for two main reasons.
First, conventional cellular networks are equipped with downtilted antennas to serve \acp{UE} on the ground. Consequently, the ground to air link is likely capacity limited or prone to low connectivity at specific areas or heights \cite{geraci2018understanding}. Second, when \acp{UAV} are deployed in unseen environments, the unavailability of knowledge about the environment increases the complexity of the path design.

\subsection{State of the art}
Prior solutions to the \ac{UAV} path optimization problem usually use conventional optimization techniques. Works in \cite{zhang2018cellularTraj_TransComm,chowdhury2020_Traje_BHConstraint,zhang2019trajectory, mardani2019communication} discuss graph-search methods, whereas a dynamic programming approach is used in \cite{bulut2018trajectory}. These approaches reformulate the corresponding non-convex path planning optimization problems in a more tractable form that suffers from poor scalability and is based on simplified assumptions on the antenna and propagation models. 

The above issues can be circumvented by exploiting detailed information on the propagation channels in a given geographical area, such as radio maps.
The work in \cite{zhang2019radio} utilizes a radio map of the environment to find the shortest path using graph algorithms. The radio map is assumed known as a priori.
In \cite{esrafilian2018learning}, the authors first reconstruct the radio map of the area to estimate the channel parameters. Then, the path is optimized to maximize the data collected from the ground. Similarly, in \cite{mardani2019communication,yang2019connectivity,pollin2019cellularCoverage}, the \ac{UAV} uses a coverage map that provides accurate locations of coverage holes in the network to maintain effective communication with the ground during the flight while moving from a starting to a final position in the shortest amount of time.
$A^{\ast}$ is thus applied for finding the shortest (or approximately shortest) path in a much shorter computation time than canonical Dijkstra’s algorithm, by considering a smaller search subspace.

The above works show that the availability of radio or coverage maps makes algorithms like $A^*$ attractive for \ac{UAV} path design problems. However, the assumption of full map availability is generally impractical since radio and coverage maps need to be estimated by collecting beforehand many radio measurements in a specific environment \cite{esrafilian2019_trajectory_3dMap_SINR}. Notably, using a $A^*$ algorithm for \ac{UAV} path design would require the availability of coverage maps for each \ac{UAV} potential and allowed height in the area where the \ac{UAV} is planned to fly. This is hard and expensive to be reached in practice. 

In \ac{RL} algorithms, such as the one proposed in this paper, prior knowledge of the environment like radio and coverage maps is not required. \ac{RL} algorithms can learn the environment and autonomously determine the optimal path through UAV-environment interactions and only with UAV's measured data, such as the received signal power.
For this reason, applying \ac{RL} algorithms in the \ac{UAV} path design has received increasing interest.
In \cite{khamidehi2020double}, a Q-learning path algorithm is proposed to design \ac{UAV} path. A \ac{UAV} interacts with the environment collecting positive or negative feedback. The algorithm considers the continuous and total outage during the \ac{UAV} path to prevent the \ac{UAV} from losing communication with the ground. 
However, when the size of the considered flying environment increases, Q-Table becomes too large and tabular methods such as Q-learning don't represent an efficient solution.
It is thus beneficial for \ac{UAV} path planning to use \ac{DRL} methods that combine \ac{RL} with \ac{DNN} to address
more challenging tasks.
In \cite{zeng2020simultaneous}, the authors study the use of \ac{DQN} to minimize the weighted sum of the \ac{UAV} mission completion time and the communication outage duration.
In \cite{chen2021joint_Augmented}, the connectivity aware path is proposed in a similar \ac{DQN} fashion, but it also includes the optimal selection of the ground \ac{BS} transmitter.

One major issue of a model-free approach in \ac{UAV} communication, such as \ac{DQN}, is the need for a relatively high number of learning trials to converge. 
During the initial training, the algorithm performance is poor and improves only when enough information about the scenario environment is collected. 
However, this lengthy training is equivalent to thousands of flights where the reliability of the ground to air link is not guaranteed and it is costly, due to the \ac{UAV}'s onboard battery and energy waste.
Preliminary works have investigated methods to improve the learning efficiency.
Using a model-based \ac{RL}, the work in \cite{zeng2020simultaneous} uses the measurements collected during the flights in the training to build a radio map of the environment.
The radio map is exploited to generate simulated UAV trajectories and predict their corresponding outage duration.
In \cite{khamidehi2020federated}, the radio map of the environment is built in a distributed fashion using \ac{FL} through the collaboration of multiple \acp{UAV}.
The joint flight and connectivity optimization problem is then solved collectively.
The work in \cite{chen2021joint_Augmented} reuses past successful trajectories to imitate the same behavior and achieve faster convergence.

\subsection{Contribution}
The above mentioned solutions contribute to reducing the algorithms' execution time but still fail to generalize when applied to different unseen scenarios. In fact, these approaches are tailored for a single environment only or used to build a radio map of the environment [17-18]. This affects the ability of the UAV to make good decisions when facing an unseen environment. As a consequence, the agent would require to re-run the lengthy training process for any new environment faced by the \ac{UAV}.

For these reasons, we believe that, to make \ac{DRL} based solutions attractive for \ac{UAV} connectivity aware path design in real scenarios, there is a need for a framework that can significantly improve the performance in unseen environments.
Motivated by this challenge, we address the \ac{UAV} connectivity aware path using a \ac{TL} approach.
\ac{TL} is the process of utilizing knowledge gained from other tasks, or prior knowledge, to benefit the target task's learning process.
The core idea of our paper is to transfer the experience gained in learning to perform the proposed robust-DDQN path design in a old domain to help improving learning performance of the proposed DDQN path design in a new domain.
Our method for transfer learning translates advice, or preferences, from a teacher path policy learned in an old domain $D_1$ at $f_1$ into a new domain $D_2$ at $f_2$. Since future wireless networks will support the sub-6 GHz and mmWave frequency ranges \cite{semiari2019integrated_mmWave_sub6_WiRCOMM_MAG}, we believe that a different frequency band represents an interesting and practical use case of unseen environment in UAV path design. 
Our approach hinges on a Lyapunov method in the search for a robust teacher policy that can effectively guarantee the connectivity constraint satisfaction during training.
To test our \ac{TL} approach in a challenging scenario, we consider sub-6 GHz and \ac{mmWave} frequency bands, which have different propagation characteristics (blockage sensitivity and scattering loss) and bandwidth availability.
While other papers focused on exploiting the correlation between these two frequencies \cite{alrabeiah2020DL_BEamPredict_usingsub6_TCOMM}, we exploit that \ac{TL} approaches are suitable for equivalent or different domains \cite{zhu2020transfer}. To demonstrate the generality of our approach we have considered different blockage scenarios corresponding to the Urban, Dense Urban and High Rise environments.

To better highlight the contribution of this paper, Table \ref{tab:Related_Works} presents a comparison of this work with different works in the literature. A systematic search was implemented to identify the most important related works in the connectivity aware design. The research is restricted to journal and conference papers only and keywords such as \ac{UAV}, connectivity and disconnectivity constraint, path and prior knowledge.
It can be noted that the connectivity outage constraint is formulated in different forms to cater for different \ac{UAV} application scenarios flexibly. 
We propose a framework that can solve the communication-aware trajectory problem efficiently without the assumptions of coverage maps while, at the same time, representing a robust teacher policy to improve the training in new environments through \ac{TL}.
To the authors' best knowledge, while \ac{TL} is becoming a crucial topic in \ac{DRL} and various domains \cite{zhang2020trajectoryTransferLearning, azab2020dynamic}, this is one of the first times Teacher Advice \ac{TL} is combined with a Lyapunov approach and applied to the \ac{UAV} connectivity aware path design. The \ac{TL} method allows us to create and incorporate prior knowledge in our \ac{DRL} solution without performing expensive measurement campaigns, speed up the learning process, and optimally solve the optimization problem.

\begingroup
\begin{table*}[t]
\centering
\caption {Comparisons between Related Studies on UAV path Optimization with Connectivity Constraint, where SNR is Signal to Noise Ratio, PER stands for Priority Experience Replay, and TD is Temporal Difference.} 
\begin{tabular}{clllll}
\hline
Ref.  &  Connectivity Constraint &   UAV Role   &   Prior Knowledge        &   Technique          \\
\hline
\cite{zhang2018cellularTraj_TransComm} & Minimum Target SNR  &   UAV-UE &  \ding{55}   & Dijkstra Algorithm \\
\cite{chowdhury2020_Traje_BHConstraint}  & Backhaul Constraint - no Minimum Rate &   UAV-BS &  \ding{55}     &  Dijkstra Algorithm \\
\cite{zhang2019trajectory} &  Maximum Outage Duration  &   UAV-UE & \ding{55}   & Graph Theory, Convex Optim. \\
\cite{mardani2019communication} &  Minimum Throughput  &   UAV-UE & Throughput Map  & $A^*$ Algorithm \\
\cite{bulut2018trajectory} & Maximum Continuous Disconnectivity Time & UAV-UE &  \ding{55}      & Dynamic Programming \\
\cite{zhang2019radio} & Minimum Target SNR   &    UAV-UE   & Radio Map &  Dijkstra Algorithm\\
\cite{esrafilian2018learning} & Minimum Target SNR &   UAV-UE &  Radio Map    & Dynamic Programming\\
\cite{yang2019connectivity} & Connectivity Outage Ratio and Duration  &   UAV-UE &  Coverage Maps    &  Graph Search Method    \\
\cite{pollin2019cellularCoverage} & Minimum Target SNR &   UAV-UE   & Coverage Map & $A^*$ Algorithm\\
\cite{esrafilian2019_trajectory_3dMap_SINR} & Minimum Target SNR  &   UAV-UE &  Radio Map    & Graph Search Method \\
\cite{khamidehi2020double} &  Maximum Continuous and Total Disconnectivity Time   &     UAV-UE    &   \ding{55}  &    Double Q-Learning     \\
\cite{zeng2020simultaneous}  & Total Disconnectivity Time   &   UAV-UE &  \ding{55}   &  Model-based DQN (Dyna)    \\
\cite{chen2021joint_Augmented} &  Maximum Continuous Disconnectivity Time &   UAV-UE &  Radio Map   &  DRL    \\
\cite{khamidehi2020federated} & Maximum Continuous Connectivity Outage &    UAV-UE &  Radio Map    &    Federated Learning \\
\cite{zhang2020trajectoryTransferLearning} &  \ding{55}   &   UAV-BS  &  Environ. Model  &   DDQN, TL    \\ 
\cite{qiu2020placement} &   Minimum Target SNR at UE   &    UAV-BS  &  Coverage Bitmap    &   PER DRL    \\ 
\cite{zeng2019path}  & Disconnection Duration &  UAV-UE  &   \ding{55}  & TD Learning \\

\cite{chen2020trajectory}  &   Connectivity Outage Ratio     &  UAV-UE  &  \ding{55}  &   Dijkstra with Intersection      \\  
\cite{chapnevis2021collaborative} &  Total Connectivity Outage Time   &  UAV-UE  &  \ding{55}  &   Dijkstra with Intersection  \\ 
\cite{wang2021learning} & UAV Disconnection Duration   &  UAV-UE  &  \ding{55}  &   Decentralized DRL \\
\cite{gao2021cellular} & UAV Disconnection Duration   &  UAV-UE  &  \ding{55}  &  DRL \\
\cite{liu2019Q_Learning} & Minimum Target SNR at UE   &  UAV-BS  &  \ding{55}  &  Q-Learning \\
\cite{khamidehi2021trajectory} &  Backhaul Constraint - Minimum Rate   &  UAV-BS  &  Channel Gain  & Interior Method \\
\cite{cherif2021CargoUAV} & Disconnectivity Rate  &  UAV-cargo  &  Connectivity Heatmap  &  Dynamic Programming\\
\cite{fontanesi2021deep} & Total Radio Failures  &  UAV-UE  &  \ding{55} &  DDQN\\
This Work &  Outage on Ground to Air Link  & UAV-UE, UAV-BS &  Teacher Policy  &  Lyapunov robust-DDQN \\
\hline
\end{tabular}
\label{tab:Related_Works}
\end{table*}
\endgroup
The contribution of this paper can be listed as follows:
\begin{itemize}
    \item First, we formulate a 3-D UAV path problem under ground to air link connectivity outage constraint as \ac{CMDP}.
    \item Thus, we propose a Lyapunov approach to solve the \ac{CMDP} and obtain a strategy that ensures the \ac{UAV} reaches the destination while respecting the connectivity outage constraint at all times.
    We then develop a robust-\ac{DDQN} based algorithm to learn an optimal policy at $f_1$. 
    \item  Utilizing the concepts of teacher advice and \ac{TL}, we present a novel algorithm that uses the derived trained policy as a teacher policy at sub-6 GHz to guide the exploration process at \ac{mmWave} and reduce the training time. 
    \item We first demonstrate the efficiency of the robust-\ac{DDQN} comparing its performance to a benchmark conventional Dueling \ac{DDQN}. At sub-6 GHz frequency band, we show that our approach can better explore the environment and achieve higher mission success. 
    \item Finally, we also evaluate the proposed teacher advice and \ac{TL} strategy in terms of the percentage of successful missions. Results show that using a teacher policy trained at sub-6 GHz frequency band significantly speeds up the learning process at \ac{mmWave} than starting the training from scratch. Moreover, the robust-\ac{DDQN} results in a better teacher policy than the state of the art Dueling \ac{DDQN}.
\end{itemize}

The system model and the problem formulation are presented in Section \ref{SystemModel}.
In Section \ref{SafeTeacherPolicy}, we transform the problem into a \ac{CMDP} and propose a robust-\ac{DDQN}-based trajecotry design algorithm to play as teacher for \ac{TL}.
The \ac{TL} approach is presented in Section \ref{TransferAdivce} while the Numerical Results are in Section \ref{NumericalResults}. Finally, Section \ref{Conclusion} concludes the paper.

\section{System Model and Problem Formulation}\label{SystemModel}
\begin{table*}[t]
 \centering
 \setlength{\tabcolsep}{3pt}
 \caption{List of Notations and Symbols Summary}\label{tab:Notation}
    \begin{tabular}{|p{55pt}|p{150pt}||p{55pt}|p{150pt}|}
        \hline
        \textbf{Notation} &  \textbf{Description} & \textbf{Notation} & \textbf{Description}\\
        \hline
        $X$ & Area of Interest &  $\Pi$ & Set of markov stationary policies\\
        $\mathcal{G}$ & The set of GBSs  & $\mathcal{C}_{\pi}()$ & Expected cumulative cost function\\  
        $V\textsubscript{max}$  & Maximum UAV speed & $\mathcal{D}_{\pi}()$ & Expected cumulative constraint function\\
        $N$ & Path duration & $\pi_B$ & Baseline policy\\
        $\overline{N}$ & Maximum UAV mission duration & $L$ & Lyapunov function\\
        $\mathbf{q}_I$ & Path starting location & $T_{\pi, h}$ & Bellman operator\\
        $\mathbf{q}_F$ & Path final location & $\epsilon$, $\hat{\epsilon}$ & Auxiliary constraint\\ 
        $q_n$ & UAV Discrete path step $n$ & $F_L$ & Set of robust policies \\
        $\mathbf{b_{m}}$  & m-th Ground BS coordinate & $L_{\epsilon}$ & Approximated Lyapunov function\\ 
        $h\textsubscript{BS}$  & BS height & $Q_D(s,a,\theta)$ & Constraint value network\\ 
        $d_{m,n}$  & Distance between the $m_{th}$ GBS and the UAV & $Q_L(q_n,a)$ & Lyapunov value function\\ 
        $L()$ & Path loss model & $Q_C(s,a,\theta)$ &  Cost value network\\ 
        $\alpha_L$, $\alpha_{NL}$ & Path loss exponents for LoS and NLoS & $Q_T(q_n,a)$ &  Stopping time value network \\ 
        $X_{L}$, $X_{NL}$ & LoS, NLoS intercepts & $p_{c}$, $p_{d}$ & Samples priority \\  
        $\phi_1$, $\phi_2$ & Antenna tilt at $f_1$, $f_2$ & $\delta$ & TD-error\\  
        $P_{TX}$ & Transmit power of ground BS  &  $\pi_T$ & Teacher policy \\ 
        $\sigma^2_n$  & Thermal noise power & $H$ & Prioritized replay memory\\
        $h$, $m_v$ & Fading, Nakagami fading parameter &   B  & Minibatch\\
        $\gamma_{m,n},\bar{\gamma}$  & SINR, SINR threshold & $\Sigma$, $\Upsilon$ & Known, Unknown space\\ 
        $S_O$ & Subset of outage Regions &  $D_1$, $D_2$  &  Old domain, new domain\\
        $F (\mathbf{q}_n)$ & Radio failure indicator & $C$, $Z$ & Size of known space memory, size\\ 
        $d_{O}$ & Connectivity outage constraint & $\Theta$, $\Lambda$ & Density threshold, Risk function\\ 
        $d_{th}$ & Max tolerated radio failures & $\pi_2$ & Policy in new domain\\ 
        \hline
    \end{tabular}
\end{table*}
We consider a set, $\mathcal{G}$, of $B$ cellular \acp{BS} providing downlink wireless service in a geographical area of interest $X \in \mathbb{R}^3$. 
\acp{UAV} can be deployed to reach an area of interest, $F \in X$, as \acp{UAV}-\ac{UE} for delivery of supplies, or \acp{UAV} as \ac{BS} or \ac{RRH} to provide service to a demand hotspot \cite{chowdhury2020_Traje_BHConstraint}.
We assume that the path of the \ac{UAV} starts from a random starting position $\mathbf{q}_I = [x_0 , y_0 , h_0] \in X, \notin F$, and ends in a final predetermined position $\mathbf{q}_F = [x_F , y_F , h_F ]^T \in F$ for a duration $T$.
For the convenience of illustration, we divide the finite \ac{UAV} mission completion time $T$ into a sequence of discrete time instances $t_1,t_2,...t_{\omega}$ such that $T=\omega \Delta_T$ and $\vert t_n-t_{n-1} \vert \leq \Delta_T$.
The \ac{UAV} path can be thus approximated by the sequence $\{ q_n\}_{n=1}^{\omega}$ where each step point at instant $n$ is thus described by its discrete coordinates $\mathbf{q_{n}}= [x_{n}, y_{n}, h_{n}]$.
The location and the transmit power $P_{BS}$ of the ground \acp{BS} can be assumed as known. We also assume that all ground \acp{BS} have equal altitude $h_{BS}$.
Each \ac{BS} and the \ac{UAV} can operate at $f_1$ and $f_2$ but we assume that data transmission occurs in a single frequency band at a time.

Let $\mathbf{b_{m}}= [x_{m}, y_{m}, h_{BS}]$ the coordinate of the $m$-th ground \ac{BS} in a three-dimensional coordinate system, the distance between the \ac{UAV} and the $m$-th ground \ac{BS} at step $n$ is given by:
\begin{equation}\label{eq:2D_distance}
    d_{m,n} = \norm{\mathbf{q_{n}}-\mathbf{b_{m}}}, \;\;\; m \in \mathcal{G}.
\end{equation}

\begin{figure}[t]
\centering
\includegraphics[width=0.9\columnwidth]{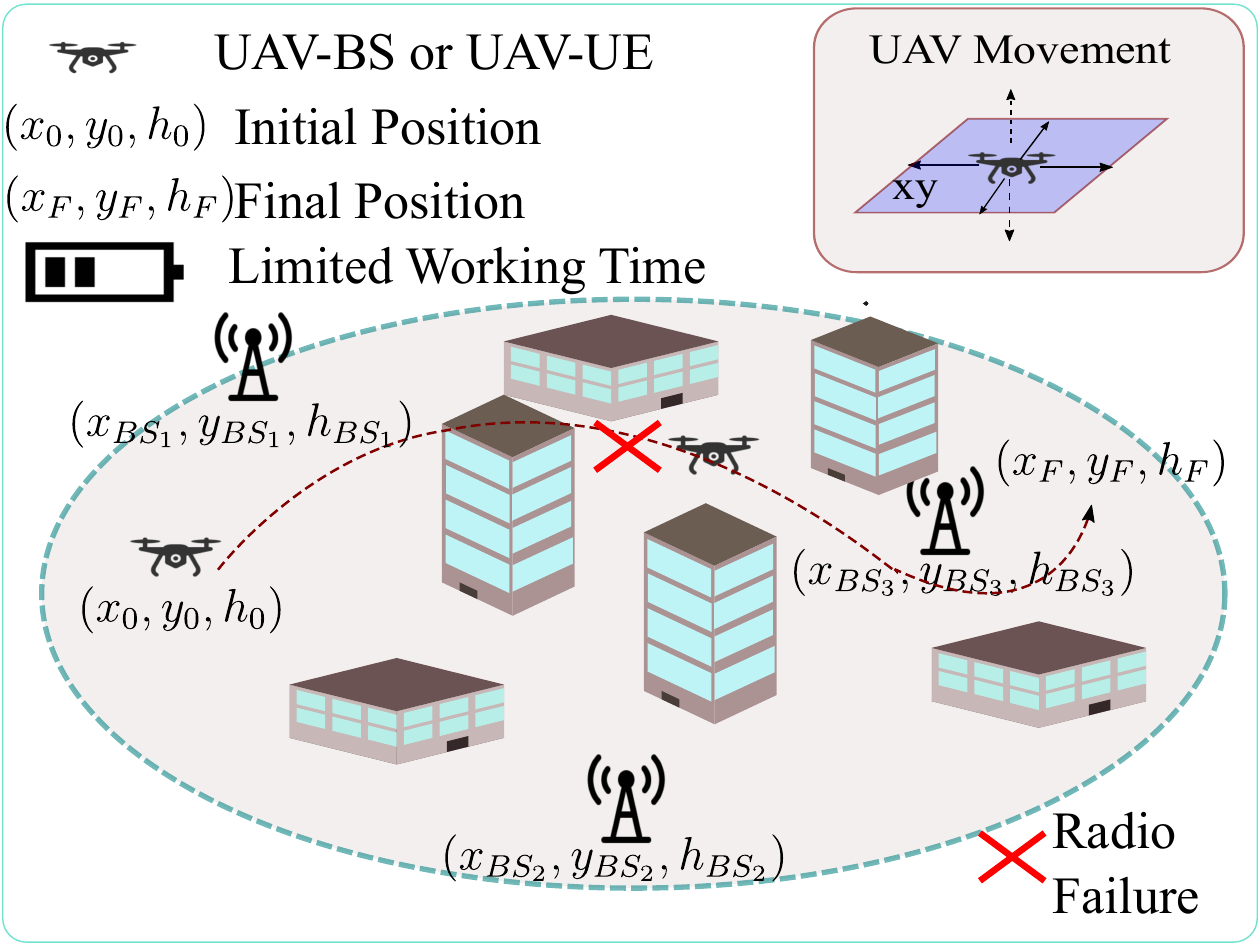} 
\caption{UAV is flying in an urban environment where the ground \ac{BS}-\ac{UAV} link might be occasionally blocked by buildings based on the building distribution and \ac{UAV} height, leading to radio failures.}
\label{fig:Trajectory_Challenge}
\end{figure}

Next, we describe the channel model and formulate the problem.

\subsection{Ground-to-Air Channel Model} \label{ChannelModel}
We consider wireless ground-to-air channel ground \ac{BS}-\ac{UAV} characterized by deterministic large scale path loss and random small-scale fading.
We consider a generic urban environment where the ground \ac{BS}-\ac{UAV} link might be occasionally blocked by obstacles and buildings based on the building distribution in $X$ and \ac{UAV} height. 
In order to present the results more generically, the path loss at $f_1$ and $f_2$ between the $m$th ground \ac{BS} and the UAV can be modeled to take into account the \ac{LoS} and \ac{NLoS} case as for \cite{fontanesi2020outage}:

\begin{equation}\label{eq:pathLoss_f2}
 L(d)=
    \begin{cases} 
       X_L d_{m,n}^{-\alpha_L};\\
       X_{NL} d_{m,n}^{-\alpha_{NL}};
    \end{cases}
\end{equation}
where $d_{m,n}$ is the ground \ac{BS}-\ac{UAV} distance as for \eqref{eq:2D_distance}, and parameters $\alpha_L$, $\alpha_{NL}$ and $X_L$, $X_{NL}$ represent, respectively, the path loss exponent for \ac{LoS}/\ac{NLoS} and the path loss at 1 meter distance.
To capture the \ac{LoS} and \ac{NLoS} effect at sub-6 GHz, we model the small scale fading power as Rician for the \ac{LoS} and as Rayleigh for the \ac{NLoS} link \cite{zeng2020simultaneous}.
At \ac{mmWave}, we model the small scale fading power $h_{0,i}^2$ with $i$ $\{LoS, NLoS\}$ as a Nakagami-\textit{$m_v$} fading model \cite{fontanesi2020outage}. 
Accordingly, the fading power at \ac{mmWave} follows a Gamma distribution with $\E[h_0^2] = 1$.

\subsubsection{Antenna Model}\label{AntennaModel}
We adopt the three-sectors antenna model as characterized by \ac{3GPP} specification \cite{3GPPStudy_LTE}.
Similar to \cite{rebato2018study}, we consider that each sector is separated by $\ang{120}$ and equipped with a vertical $N$-element \ac{ULA} tilted with angle $\phi_1$ at $f_1$ and a \ac{UPA} $N \times N$ tilted with angle $\phi_2$ at $f_2$. 
The dB gain experienced by a ray with elevation and azimuth angle pair $\theta, \phi$ due to the effect of the element radiation pattern can be expressed as:
\begin{equation}\label{eq:antennaGain}
    A_E^{3GPP} (\theta, \phi) = G_{max}-\min\{-[A_{E,V}(\theta)+A_{E,H}(\phi)], A_m \}
\end{equation}
where $ G_{max} = 8$ dBi is the maximum directional gain of the antenna element.
The \ac{3GPP} element radiation pattern of each single antenna element is composed of horizontal and vertical radiation patterns $A_{E,H}(\phi)$ $A_{E,V}(\theta)$.
Specifically, this last pattern $A_{E,V} (\theta)$ is obtained as
\begin{equation}
    A_{E,V} (\theta) = -\min \bigg\{ 12\bigg(\frac{\theta -90\degree}{\theta_{3 dB}} \bigg)^2, SLA_V \bigg\}
\end{equation}
where $\theta_{3dB} = 65\degree$ is the vertical beamwidth, and $SLA_V = 30$ dB is the side-lobe level limit.
Similarly, the horizontal pattern is computed as
\begin{equation}
     A_{E,H} (\phi) = -\min \bigg\{ 12\bigg(\frac{\phi}{\phi_{3dB}} \bigg)^2, A_m \bigg\}
\end{equation}
where $\phi_{3dB} = 65\degree$ and $A_{m} = 30$ dB is the front-back ratio. 
The relationship between the array radiation pattern and a single pattern is defined as $A_A(\theta, \phi, n) = A_{E}(\theta, \phi) + AF(\theta, \phi, n) $, where $n$ is the number of antenna elements and $AF(\theta, \phi, n)$ is the array factor. $AF(\theta, \phi, n)$ is given in \cite{rebato2018study} as:
\begin{equation}
    AF(\theta, \phi, n) = 10\log_{10} \bigg[1+ \rho \bigg( \big|\textbf{a} \cdot \textbf{w}^T \big|  \bigg)  \bigg]
\end{equation}
where $\rho$ represents the correlation coefficient set to unity, $\textbf{a}$ is the amplitude vector and $\textbf{w}$ is the beamforming vector. The definition of $\textbf{w}$ can be found in \cite{chen2021joint_Augmented,rebato2018study} and is omitted here. As in \cite{chen2021joint_Augmented,rebato2018study}, we consider an equal and fixed amplitude for the antenna elements. As a consequence, $\textbf{a}$ is set to $(\frac{1}{\sqrt{n}})$ and we leave the integration of optimized weights in beamforming techniques for future investigation.

Finally, we assume the \ac{UAV} is equipped with a conventional isotropic antenna of unitary gain in any direction to mantain low complexity and cost.

For simplicity, but without loss of generality, we focus on one typical \ac{UAV}.
At any time step during the \ac{UAV}'s mission, the \ac{UAV} associates with one ground \ac{BS}'s sector only, following a maximum \ac{SINR} scheme \cite{fontanesi2020outage}.
The maximum instantaneous \ac{SINR} received at the omnidirectional \ac{UAV} from the attached $m$-th ground \ac{BS} can be defined as:
\begin{equation}\label{eq:snr}
    \gamma_{m,n}=\frac{ P_{BS}L(d_{m,n}) h_0^2(d_{m,n}) g_{m^*,j^*}}{\sigma^2+I_t},
\end{equation}
where $L()$ is the path loss, the random variable $h$ accounts for the fading, and $I_t$ the interference associated with the non-attached \acp{BS}. 
The term 
\[g_{m^*,j^*} = \displaystyle \argmax_{m \in \mathcal{G}, j \in \mathcal{J}} g_{m,j}\]
is the antenna gain from the $j$-th sector of the $m$-th ground \ac{BS} and $j \in \mathcal{J} =\{1,2,3\}$ denotes the set of sectors.

For a given \ac{SINR} threshold $\bar{\gamma}$, an outage occurs at step $n$ if at the \ac{UAV} the condition $\gamma_{m,n} \leq \bar{\gamma} $ is not satisfied.
The resulting outage probability can be denoted as 
\begin{equation}\label{eq:outage}
P_{outage}(\mathbf{q}_n, m) = Pr(\gamma_{m,n} < \bar{\gamma} ),   
\end{equation}
where $Pr$ is the probability of the event taken with respect of the randomness of the fading.
Note that the value of $\Delta_T$ can be considered small enough to satisfy $\Delta_T V << h_{n}$, and in the generic flying step $n$, the \ac{UAV} can be considered stationary \cite{zeng2020simultaneous, zhang2021SAFEDQN}.
Let us define a radio failure indicator on the ground to air link as
\begin{equation}\label{eq:radiofailure}
F (\mathbf{q}_n) =
    \begin{cases}
    1, & \mbox{if } P_{outage}(\mathbf{q}_n, m) \geq \bar{P}_{th}\\
    0, & \mbox{otherwise}.
\end{cases}
\end{equation}  
Thus, we can introduce $S_O$ as the subset of outage regions where \eqref{eq:radiofailure} holds true. Then, for an arbitrary outage probability threshold $\bar{P}_{th}$ and a given path $q_n$ with $\mathbf{q}_I \notin S_O$, the connectivity outage constraint $d_{O}$ can be expressed as
\begin{equation}\label{eq:SafetyCMDP}
 d_{O} = \sum_{n=1}^{N} F(\mathbf{q}_{n}).
\end{equation}

\subsection{Problem Formulation}

We would like the \ac{UAV} to reach the destination in the shortest possible number of moves, while keeping the outage events lower than $d_{th}$.
The UAV’s velocity is limited to its maximum speed.
We consider that during its mission, the UAV moves at constant $V=V\textsubscript{max}$.
This assumption of constant maximum speed makes the mathematical modeling more tractable as the variable speed will have control and/or aerodynamic related reasons which are out of our control and scope of work. In addition, using UAV's maximum speed allows the UAV to reach the destination in the minimum path steps. The UAV maximum speed used in this paper to derive the numerical results is a realistic and aerodynamic supported maximum speed, used in several related connectivity-aware path design works \cite{chen2020trajectory,zeng2020simultaneous, khamidehi2020federated}.
Thus, the mission variable $T$ can be expressed as $T = \frac{\sum_{n=1}^\omega \norm{\mathbf{q}_n-\mathbf{q_{n-1}}}}{V_{Max}}$ and optimization problem can be formulated as in \eqref{eq:Opt_problem1_discrete}.
\begin{subequations}\label{eq:Opt_problem1_discrete}
    \begin{align}
    \min_{\omega} \quad & \sum_{n=1}^\omega \norm{\mathbf{q}_n-\mathbf{q_{n-1}}}\\ \label{eq:cond1_Opt_problem1_discrete}
    \textrm{s.t.} \quad & \mathbf{q}_0 = \mathbf{q}_I,\mathbf{q}_N = \mathbf{q}_F\\ \label{eq:cond2_Opt_problem1_discrete}
     & h_n > h_{BS}\\\label{eq:cond3_Opt_problem1_discrete}
     & \norm{\mathbf{q}_n -\mathbf{q}_{n-1}} \leq \Delta_T V\textsubscript{max},\\ \label{eq:cond4_Opt_problem1_discrete}
    & d_{O} \leq d_{th},\\ \label{eq:cond5_Opt_problem1_discrete}
    & \omega \leq \bar{N}.
    \end{align}
    \end{subequations}
Variable $\omega$ represents the mission completion steps. 
The constraint \eqref{eq:cond1_Opt_problem1_discrete} guarantees the initial and final points, \eqref{eq:cond2_Opt_problem1_discrete} prevents the collision between the \ac{UAV} and the ground \acp{BS}.
\eqref{eq:cond3_Opt_problem1_discrete} constraints on the UAV's maximum speed.
The connectivity constraint $d_{O}$ must not exceed a predefined threshold $d_{th}$. For this reason, \eqref{eq:cond4_Opt_problem1_discrete} defines the connectivity outage duration tolerance. Constant $\overline{N}$ in \eqref{eq:cond5_Opt_problem1_discrete} is the upper bound on the \ac{UAV} steps to take into account the limited \ac{UAV} endurance.
In our design, $d_{th}$ is not fixed but can be tuned to suit different application scenarios.
Longer paths may help the \ac{UAV} avoid $S_O$ areas and satisfy stringent values of $d_{th}$.
In scenarios where the \ac{UAV} is deployed for timing intervention a higher $d_{th}$ might be tolerated to achieve shorter paths.
Although \ac{UAV}-\ac{UE} and \ac{UAV}-\ac{BS} cases generally have different design problems, the above connectivity constraint path optimization applies to both the scenarios from the ground to air link point of view.

The connectivity-aware problem \eqref{eq:Opt_problem1_discrete} is a non-convex optimization problem that is generally intractable to solve via conventional optimization techniques. The Proof of NP-hardness of a similar path design problem can be found in \cite{chen2020trajectory} and omitted here. In addition, a closed-form expression of the outage probability \eqref{eq:outage} used to compute \eqref{eq:cond4_Opt_problem1_discrete} is highly dependent on the network topology, channel fading and antenna gain. In our previous work \cite{fontanesi2020outage}, taking into account the channel characteristics at $f_1$ and $f_2$, we have investigated a stochastic geometry approach to deduce a tractable form of $P_{outage}$. 
However, statistical approaches provide useful insights on the average performance of the network but they don't capture the actual complexity of the local environment where the \acp{UAV} are deployed.

\ac{RL} approaches, that interact iteratively with the environment, circumvent these issues solving the path optimization problem using the power measurements at the \ac{UAV} in a certain time step. While \ac{RL} algorithms for the design of UAV connectivity aware path have been proposed already in literature (Table \ref{tab:Related_Works}), this paper aims to propose a novel \ac{TL} approach to improve the efficiency of \ac{DQN} for \ac{UAV} path design.
More specifically, adopting the dual band system model described in Section \ref{SystemModel}, we focus on two fundamental problems: (i) how can a robust policy derived at $f_1$ be used to infer the path at $f_2$, (ii) what is the best algorithm solution of \eqref{eq:Opt_problem1_discrete} at $f_1$ to act as teacher for $f_2$, solving (i).

Next, to determine a set of suitable policies to the connectivity aware problem, we propose a Lyapunov approach method to the \ac{UAV} path design.
Finally, based on the Lyapunov approach, we develop a teacher robust-\ac{DDQN} algorithm.

\section{Creation of a robust Teacher Policy}\label{SafeTeacherPolicy}
The position of \ac{UAV} at step ${q_{n+1}}$ depends on the position and moving direction chosen by the agent at step $q_n$. Hence, the UAV's flight process can be regarded as a discrete-time \ac{CMDP}, an extension of the \ac{MDP} framework that suits optimization problems as \eqref{eq:Opt_problem1_discrete}, where agents optimize one objective while satisfying cost constraints. 
Note that we assume the \ac{UAV} to be controlled by a dedicated agent and the terms agent and \ac{UAV} will be used hereafter interchangeably.
\subsection{Problem Formulation as CMDP} \label{ProblemFormulationCMDP}
Each \ac{CMDP} consists of a 7-tuple $\langle \mathcal{S}, \mathcal{A}, \mathcal{P}, c, d, \alpha, d_{th} \rangle$.
In this work $\mathcal{S}$ is the space state consisting of the \ac{UAV} positions within the feasible flying region with single state $\mathbf{q}_n$. 
The agent's action corresponds to one of the \ac{UAV}'s flying directions, that together represent the action space $\mathcal{A}$. The UAV moves in a custom environment consisting of area $X$ covered by the dual band network.
The immediate cost function is modeled as $c(\mathbf{q}_n, a_n) = -1 -\norm{\mathbf{q_{n}}-\mathbf{q_{F}}}$. The first term penalizes the UAV for each move and accounts for \ac{UAV} battery usage. At the same time, the second cost term measures the relative distance to the destination and encourages the \ac{UAV} to complete its mission in the shortest possible number of moves.
We define the immediate constraint cost $d(\mathbf{q}_n)$ as the radio failure indicator $F(\mathbf{q}_n)$ in \eqref{eq:radiofailure}, $d_{th}$ is an upper bound on the expected cumulative constraint cost.
Lastly, variable $\alpha \in [0,1]$ in $\langle \mathcal{S}, \mathcal{A}, \mathcal{P}, c, d, \alpha, d_{th} \rangle$ represents the discount factor. 

Under this framework, at any step $n$, the \ac{UAV} moves from $\mathbf{q_{n}}$ to $\mathbf{q_{n+1}}$ during step length $\Delta_T$ at speed $V\textsubscript{max}$ based on the action $a_n$, selected according the current policy $\pi$. We define a policy $\pi(a \!\!\mid \!\!\mathbf{q}_n)$ as the condition probability to take action $a$ given the state $\mathbf{q}_n$.
After taking action $a_n$, the \ac{UAV} interacts with the environment receiving the immediate cost $c(\mathbf{q}_n, a_n)$ and constraint cost $d(\mathbf{q}_n)$.  
A sequence of interactions leads to a terminal or goal state, where an episode ends.
For the computation $F(\mathbf{q}_n)$, note that $\Delta_T$ usually contains many channel coherence blocks due to small-scale fading. As a result, it can be assumed that as long as the \ac{UAV} performs signal measurements sufficiently frequently, the outage probability \eqref{eq:outage} can be evaluated by its empirical value  $\hat{P}_{outage}(\mathbf{q}_n, m) = 1/J\sum_{j=1}^{J} O(\mathbf{q}_n,m)$ where $J \gg 1 $ and $O ()$ is 1 if $\gamma_{m,n} \leq \bar{\gamma} $ and 0 otherwise \cite{zeng2020simultaneous}.

To complete the \ac{CMDP} formulation of \eqref{eq:Opt_problem1_discrete} we need to formalize the constraint \eqref{eq:cond4_Opt_problem1_discrete} that bounds the total frequency of visiting $S_O$ with a predefined threshold $d_{th}$ into a \ac{CMDP}. We rewrite \eqref{eq:cond4_Opt_problem1_discrete} using the immediate constraint notation $d(\mathbf{q}_n) =  \textbf{1}\{ \mathbf{q}_n \in S_O\}$, where \textbf{1}\{x\} denotes the indicator function so that its value is 1 if $x\in S_O$ and 0 otherwise \cite{chow2018lyapunov}. Thus constraint \eqref{eq:cond4_Opt_problem1_discrete} becomes 
\begin{equation}\label{eq:maxFailures_CMDP}
  \E [ \sum_{n=1}^{N}\textbf{1}\{ \mathbf{q}_n \in S_O\}\mid \mathbf{q}_I,\pi] \leq d_{th}.   
\end{equation}

Let us now denote $\Pi$ the set of Markov stationary policies with policy element $\pi$, such that $\Pi (\mathbf{q}_n) = \{ \pi(\cdot \!\! \mid \!\! \mathbf{q}_n) : \mathcal{S} \xrightarrow{}\mathcal{R}_{\geq 0s} : \sum_a \pi(a \!\!\mid \!\!\mathbf{q}_n) = 1\}, \forall \mathbf{q}_n \in \mathcal{S}$ follows from the stationary property. 
Given a policy $\pi \in \Pi$ that maps states to actions and an initial state $\mathbf{q}_I$, we can define the expected cumulative cost function as
\begin{equation}
 \mathcal{C}_{\pi}(\mathbf{q}_I) = \E \big[ \sum_{n=1}^{N} c(\mathbf{q}_n, a_n) \mid \mathbf{q}_I,\pi \big],
\end{equation}
and the robustness constraint function as
\begin{equation}
\mathcal{D}_{\pi}(\mathbf{q}_I) = \E \big[ \sum_{n=1}^{N} d(\mathbf{q}_n) \mid \mathbf{q}_I, \pi \big]. 
\end{equation} 
The optimization problem \eqref{eq:Opt_problem1_discrete} becomes then
\begin{subequations}\label{eq:CMDP_opt_problem}
\begin{align}
\pi^* \in \min_{\pi} \E \big[\sum_{n=1}^{N} c(\mathbf{q}_n,a_n) \mid \mathbf{q}_I,\pi \big]\\ \label{eq:cond1_CMDP_opt_problem}
\textrm{s.t.} \E \big[\sum_{n=1}^{N} d(\mathbf{q}_n) \mid \mathbf{q}_I,\pi \big] \leq d_{th}.
\end{align}
\end{subequations}
The goal of the agent is to find the optimal policy $\pi^*$ that minimizes the long term cost while satisfying the connectivity constraint.

We propose using the Lyapunov function-based method to derive a robust optimal policy $\pi^*$, solution of \eqref{eq:CMDP_opt_problem} in domain $D_1$ at $f_1$.
The rationale behind the Lyapunov approach is to find a set of robust actions that meet the condition \eqref{eq:cond1_CMDP_opt_problem} and guarantee global robustness during training.
As a consequence, it can be considered a suitable candidate as teacher in the \ac{TL} process.
From here on, for simplicity, we will refer to the robust policy $\pi_1$ in domain $D_1$ as teacher policy $\pi_T$. 
To the best of the authors' knowledge, this is the first time a Lyapunov approach is used to derive a teacher policy for transfer advice for the \ac{UAV} connectivity-aware path problem. 

\subsection{Background of the Lyapunov-Based robust Policy}\label{LyapunovPolicy}

We introduce the notation of Lyapunov function following the definition in \cite{chow2018lyapunov}: Given a baseline policy $\pi_B$, i.e. $\mathcal{D}_{\pi_B}(\mathbf{q}_I) \leq d_{th}$, a function $L: S \xrightarrow{}\mathcal{R}$ is said to be a Lyapunov function w.r.t initial state $q_I$ and constraint threshold $d_{th}$ if it satisfies the following conditions:
\begin{subequations}\label{eq:LyapunovDef}
\begin{align}\label{eq:LyapunovCondition}
    T_{\pi_B, d} [L](\mathbf{q}_n)  &\leq L(\mathbf{q}_n) \quad \forall (\mathbf{q}_n) \in \mathcal{S},\\\label{eq:SafetyCondition}
    L(\mathbf{q}_I) &\leq d_{th}, \quad\\ 
    L(\mathbf{q}_F) &= 0.
\end{align}
\end{subequations}
We denote the constraints in \eqref{eq:LyapunovCondition} as Lyapunov constraints and \eqref{eq:SafetyCondition} as robustness condition.
Term $T_{\pi, d}$ is the generic Bellman operator w.r.t a policy $\pi$ and constraint cost function $d$ 
\begin{multline}
 T_{\pi, d} [V](\mathbf{q}_n) =\\
 \sum_a \pi(a \mid \mathbf{q}_n) \big[d(\mathbf{q}_n,a) +\!\! \sum_{q_{n}' \in \mathcal{S}} P(\mathbf{q}_{n}' \mid \mathbf{q}_n,a) V(\mathbf{q}_{n}') \big],  
\end{multline}
where $\mathbf{q}_{n}'$ is next state $\mathbf{q}_{n+1} \in \mathcal{S}$ under action $a$.

Given any arbitrary Lyapunov function $L$, consider the set $F_L(\mathbf{q}_n) = \{ \pi( \cdot \mid \mathbf{q}_n) \in \Pi: T_{\pi_B, d} [L](\mathbf{q}_n) \leq L(\mathbf{q}_n) \} \; \forall \mathbf{q}_n \in \mathcal{S}$.
Given the contraction property of $T_{\pi_B, d}$ \cite{sutton2018reinforcement}, together with $L(\mathbf{q}_I) \leq d_{th}$, any policy $\pi$ in this set satisfies the robustness conditions and is a feasible solution of \eqref{eq:CMDP_opt_problem}.
This set of robust policies is defined as the L-induced policy set.
Since it is not guaranteed that the set $F_L(\mathbf{q}_n)$ contains any optimal solution of \eqref{eq:CMDP_opt_problem}, the goal of the Lyapunov approach in this paper is to formulate an appropriate function $L$, such that $F_L$ contains an optimal policy $\pi^*$ to work as $\pi_T$.
Finding an appropriate Lyapunov function may not be an easy task.
\cite{chow2018lyapunov}[Lemma 1] ensures that without loss of optimality, the Lyapunov function that satisfies the above criterion can be expressed as
\begin{equation}\label{eq:LyapunovFunction}
    L_{\pi_B,\epsilon}(\mathbf{q}_n) = \E \big[ \sum_{n=0}^{N} d(\mathbf{q}_n) + \epsilon(\mathbf{q}_n) \mid \pi_B, \mathbf{q}_n \big],
\end{equation}
in which $\epsilon(\mathbf{q}_n)$ is an auxiliary constraint. 
Thus, finding $L$ that satisfies the above condition is equivalent to perform appropriate cost-shaping with auxiliary $\epsilon$ which can be built using the method proposed in \cite{chow2018lyapunov}. This method approximates $\epsilon$ to a constant function, which is independent of state and can be computed more efficiently as
\begin{equation}\label{eq:maximizer}
    \hat{\epsilon} = \dfrac{d_{th}-D_{\pi_B}(\mathbf{q}_I)}{\E [T^* \!\!\mid\!\! \mathbf{q}_I, \pi_B]}, \mathbf{q}_n \in \mathcal{S},
\end{equation}
where $\E [T^* \!\!\mid\!\! \mathbf{q}_I, \pi_B]$ is the expected stopping time of the \ac{CMDP}.
To speed up the computation of the expected stopping time we replace the denominator of \eqref{eq:maximizer} with the upper bound $\overline{N}$, maximum number of allowed steps, leading to $ \hat{\epsilon} = \frac{1}{\overline{N}} (d_{th}-D_{\pi_B}(\mathbf{q}_I))$.
Substituting this last equation into \eqref{eq:LyapunovFunction}, the Lyapunov function becomes
\begin{equation}
    L_{\pi_B,\epsilon} (\mathbf{q}_n) = \E \big[ \sum_{n=0}^{N} d(\mathbf{q}_n) + \hat{\epsilon} \mid \pi_B,\mathbf{q}_n \big],
\end{equation}
and the set of robust policies $F_L(q_n)$ can be written as
\begin{equation}
     F_L(\mathbf{q}_n) = \{ \pi( \cdot \mid \mathbf{q}_n) \in \Pi: T_{\pi_B, d} [L_{\hat{\epsilon}}](\mathbf{q}_n) \leq L_{\pi_B,\hat{\epsilon}}(\mathbf{q}_n) \}.   
\end{equation}
The above formulation can be used to propose a robust policy and value iteration algorithm, in which the goal is to solve the \ac{LP} problem \cite{chow2018lyapunov}
\begin{multline}\label{eq:LP_Lyap}
    \pi^{*}(\cdot \mid \mathbf{q}_n) \in \argmin_{\pi \in \Pi}  \big\{ \pi (\cdot \mid \mathbf{q}_n)^T Q_C(\mathbf{q}_n,\cdot):\\
    (\pi (\cdot \mid \mathbf{q}_n)-\pi_{B} (\cdot \mid \mathbf{q}_n))^T Q_L(\mathbf{q}_n,\cdot) \leq \hat{\epsilon}  \big\}
\end{multline}   
where $Q_L(\mathbf{q}_n,a) = d(\mathbf{q}_n) + \hat{\epsilon} + \alpha \sum P(\mathbf{q}_{n}' \mid \mathbf{q}_n,a) L_{\pi_B,\hat{\epsilon}}$ is the Lyapunov function and $Q_{C} (\mathbf{q}_n,a ) = c(\mathbf{q}_n,a) + \alpha \sum_{\mathbf{q}_{n}'}P(\mathbf{q}_{n}' \mid \mathbf{q}_n,a) V_{C}$ and $V_{C} (\mathbf{q}_n) = T_{\pi_{B,c}}[V_{C}](\mathbf{q}_n)$ are the state action value function and the value function (w.r.t. the cost function c).

Since we assume that the environment in which the \ac{UAV} is flying is composed by a large and continuous state space, solving \eqref{eq:LP_Lyap} becomes numerically intractable. To address this issue, in the next section, we propose a \ac{DDQN} approach.

\subsection{Lyapunov Approach DDQN for Connectivity-Aware Path Design}\label{SafeLyapunovDQN}
In this section we use the above derived Lyapunov function to derive an optimal robust policy via \ac{DDQN}.
Using the notation of action-value function \cite{sutton2018reinforcement}, we can write the Lyapunov state-action value function $Q_L(q_n,a)$ as
\begin{equation}\label{eq:stateActionLyapunovfunction}
Q_L(\mathbf{q}_n,a) = Q_D(\mathbf{q}_n,a) + \hat{\epsilon}Q_T(\mathbf{q}_n).   
\end{equation}
where $Q_D(\mathbf{q}_n,a)$ represents the constraint state-action value function. 
The stopping time value network $Q_T(\mathbf{q}_n)$ is a function related to the number of remaining steps and discount factor, and can be computed as  $Q_T(\mathbf{q}_n)= \sum_{t=m}^{\overline{N}+1-m} \alpha^{t-m}, \forall \mathbf{q}_n \in S$.

If $\pi_B$, $Q_D(\mathbf{q}_n,a)$ and $Q_T(\mathbf{q}_n)$ are known, the auxiliary cost in \eqref{eq:maximizer} can be computed as
\begin{equation}\label{eq:epsilon_segnato}
    \epsilon'(\mathbf{q}_n) = \epsilon' = \frac{d_{th}-\pi_B(\cdot \mid s_0)^T Q_D(\mathbf{q}_n,a)}{\pi_B(\cdot \mid s_0)^T Q_T(\mathbf{q}_n)}.
\end{equation}

Finding the optimal policy $\pi^*$ through \eqref{eq:LP_Lyap} and \eqref{eq:epsilon_segnato} requires accurate calculation of $Q_D(\mathbf{q}_n,a)$, $Q_C(\mathbf{q}_n,a)$ and $\pi_B$. 
One traditional way to derive optimal action-value functions is table-based method, which requires storing and maintaining a state-action value table, one value for each state-action pair.
However, for the path design under consideration, the state-action value table would exponentially grow with the size of the flying area. 
To overcome this issue, parametric functions can be trained to approximate the state-action value. Specifically, we utilize \acp{NN} to perform function approximation.
Let $\hat{Q}_D(\mathbf{q}_n,a; \theta_D)$, $\hat{Q}_C(\mathbf{q}_n,a, \theta_C)$ be the parameterized evaluation networks with weights $\theta_D$ and $\theta_C$, then \eqref{eq:stateActionLyapunovfunction} becomes 
\begin{equation}
    Q_L(\mathbf{q}_n,a, \theta_D) = \hat{Q}_D(\mathbf{q}_n,a, \theta_D) + \hat{\epsilon}'Q_T(\mathbf{q}_n). 
\end{equation}
where $\hat{\epsilon}'$ is computed as
\begin{equation}
    \hat{\epsilon}'(\mathbf{q}_n) = \hat{\epsilon}' = \frac{d_{th}-\pi_B(\cdot \mid s_0)^T \hat{Q}_D(\mathbf{q}_n,a; \theta_D)}{\pi_B(\cdot \mid s_0)^T Q_T(\mathbf{q}_n)}.
\end{equation}
To train the networks $\hat{Q}_D$, $\hat{Q}_C$ we minimize squared error of prioritized Bellman residuals as for a loss function that can be defined as
\begin{equation}\label{eq:LossQ_C}
    L_c(\theta_C) = p_{c,n} \big(y^{c}_n - \hat{Q}_C(\mathbf{q}_n,a, \theta_C)  \big)^2,
\end{equation}
and
\begin{equation}\label{eq:LossQ_D}
    L_d(\theta_D) = p_{d,n} \big(y^{d}_n - \hat{Q}_D(\mathbf{q}_n,a, \theta_D)  \big)^2,
\end{equation}
where $p_{c,n}$ and $p_{d,n}$ are the samples priority.
In the above equations, term $y^{c}_n$ is the target cost value, expressed as
\begin{equation}\label{eq:targetValue}
 y^{c}_n = c_{n:n+N_1}+\alpha^{N_1}\pi(\cdot \mid \mathbf{q}_{n}')^T \hat{Q}_C(\mathbf{q}_{n+N_1},a^*, \theta_C^{-}),
\end{equation}
where $a^*$ is
\begin{equation}\label{eq:actionEvaluation}
a^* = \argmax \hat{Q}_C(\mathbf{q}_{n+N_1},a', \theta_C),
\end{equation}
to separate the action selection and the action evaluation as for double Q-learning technique \cite{hado2010hasselt}.
Similarly, the target $y^{d}_n$ for the constraint cost can be denoted as
\begin{equation}\label{eq:targetValue_constraint}
 y^{d}_n = d_{n:n+N_1}+\alpha^{N_1}\pi(\cdot \mid \mathbf{q}_{n}')^T \hat{Q}_D(\mathbf{q}_{n+N_1},a^*, \theta_D^{-}).
\end{equation}

In each iteration the agent takes action $a_n$ generated by current baseline policy $\pi_B$, and perform a \ac{DDQN} update, computing the loss to update the weights $\theta_C$, $\theta_D$ of networks $\hat{Q}_D$, $\hat{Q}_C$.
To derive a reasonable baseline policy $\pi_B$ for the UAV path design under study we create another \ac{DNN}. As a result, the baseline strategy action probability is approximated by the output of the \ac{DNN}, namely $\pi_B \approx \hat{\pi}(\cdot \mid \mathbf{q}_n; \theta_{\pi})$.
We train the policy network by optimizing a loss function that consists on the \ac{KL} divergence between the baseline strategy and the optimal strategy as:
\begin{equation}\label{eq:KLDivergence}
    L(\theta_{\pi}) = \E_{\mathbf{q}_n}[D_{KL}(\hat{\pi}(\cdot \mid \mathbf{q}_n;\theta_{\pi})\mid\mid \pi^*(\cdot \mid \mathbf{q}_n))].
\end{equation}

Note that in equations \eqref{eq:LossQ_C}-\eqref{eq:targetValue_constraint}, to improve the stability and convergence of our algorithm, we exploit different techniques.
Unlike the conventional Q-learning where target functions are produced by using one-step look-ahead, we use n-step lookaheads, or multi-step learning technique. 
Specifically, in the target equation \eqref{eq:targetValue}, \eqref{eq:targetValue_constraint} the truncated $N_1$-step cost and constraint cost from a given state $\mathbf{q}_n$ are defined as:
\begin{equation}\label{eq:C_Nstep}
    c_{n:n+N_1} = \sum_{i=0}^{N_1-1} \alpha^i c_{n+1+i}
\end{equation}
\begin{equation}\label{eq:D_Nstep}
 d_{n:n+N_1} = \sum_{i=0}^{N_1-1} \alpha^i d_{n+1+i}.
\end{equation}
In conventional \ac{DRL}, after executing the action, the agent stores the state-action-reward transition into a replay memory. In a second step, the agent performs the weight updates selecting a random sample of $\lvert B\rvert$ instances to break the correlation between instances \cite{kapturowski2018recurrent}.
However, sampling randomly the mini-batch $B$ may affect the convergence of the training procedure. For this reason samples can be selected according to a priority determined by their \ac{TD} error, which can be computed as $\delta_c = \{ y^{c}_j-\hat{Q}_C(\mathbf{q}_n,a, \theta_C) \}_{j=1}^{\mid B\mid}$, $\delta_d = \{ y^{d}_j-\hat{Q}_D(\mathbf{q}_n,a, \theta_D) \}_{j=1}^{\lvert B\rvert}$.
In this work, we apply a replay prioritization scheme that considers that target functions are produced by using a multi-step learning technique. Samples and TD errors are stored in a sliding window $W$ for $N_1$ transitions to enable multi-step learning. The sampling priority $p_{c,n}$ and $p_{d,n}$ in \eqref{eq:LossQ_C}, \eqref{eq:LossQ_D} are given by a weighted sum of two different components as
    \begin{equation}\label{eq:priority}
       \eta \max_i \delta_i + (1-\eta) \bar{\delta}
    \end{equation}
where in the general $\delta$ we omitted the subscript $c$ or $d$ to simplifu the notation. $\delta_c$ is used for the computation of $p_{c,n}$ and $\delta_d$ for $p_{d,n}$.
The term $\max_i \delta_i$ is the max absolute $N_1$-step TD error $\delta$ contained within the $\vert B\vert$-length sequence, $\eta$ is a tunable parameter $\in [0,1]$. The second term is the sequence mean absolute $N_1$-step TD error.
Finally, it can be noted in \eqref{eq:targetValue}, \eqref{eq:targetValue_constraint} that
$\hat{Q}_C(\mathbf{q}_{n+N_1},a^*, \theta_C^{-})$, $\hat{Q}_D(\mathbf{q}_{n+N_1},a; \theta_D^{-})$ are target networks of the evaluation networks. A target network has the identical \ac{NN} structure of the related evaluation network, but its weights $\theta_C^{-}$,$ \theta_D^{-}$ are updated only each $I$ iterations by copying the weights from the evaluation. In this way, the correlation between the target and estimated Q-values is reduced.

\begin{figure}[t]
    \centering
    \includegraphics[]{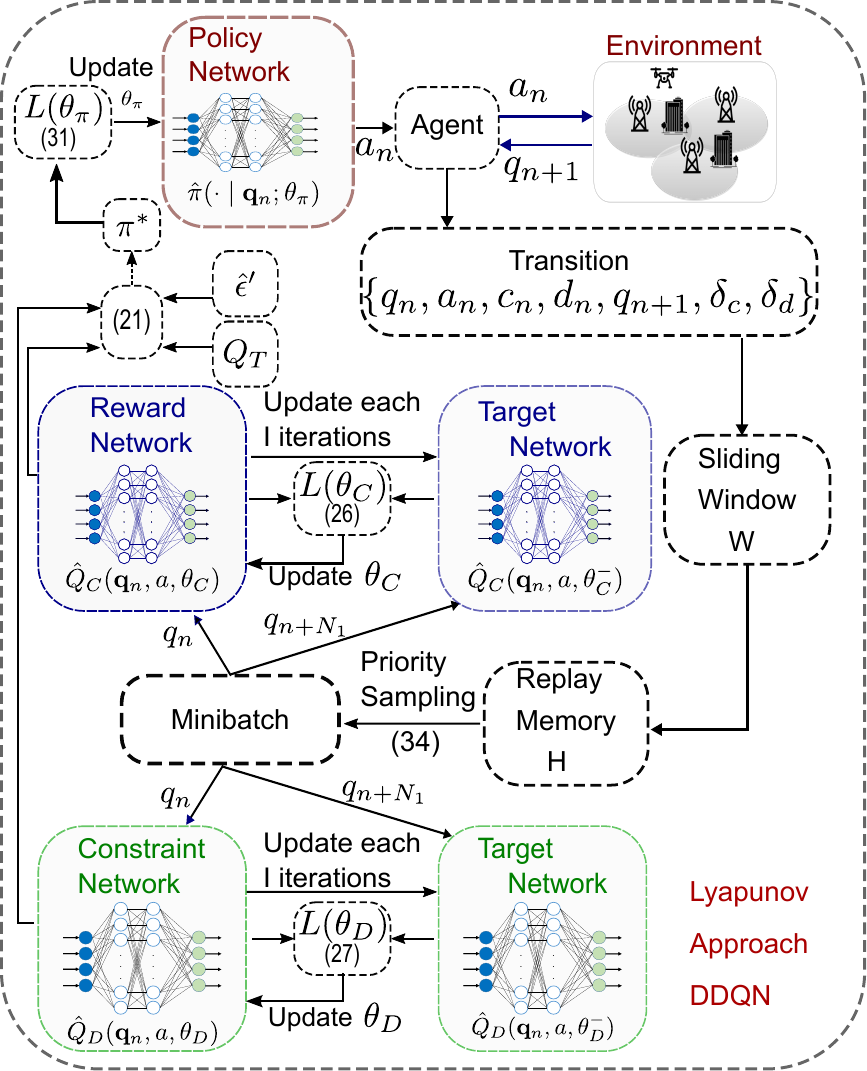}
    \caption{The Lyapunov robust-DDQN scheme proposed in this paper for connectivity-aware path design with Policy, Cost and Constraint Cost networks.}
    \label{fig:SafeDQNSCheme}
\end{figure}

The proposed algorithm to derive a robust \ac{UAV} path via Lyapunov method is summarized in \figref{fig:SafeDQNSCheme}, while Algorithm \ref{alg:SafeDDQN} presents the pseudocode. 

\begin{algorithm}[ht]
\caption{robust DDQN Algorithm for Connectivity-Aware Path}\label{alg:SafeDDQN}
\SetAlgoLined
\textbf{Initialize:} maximum number of episodes, the prioritized replay memory $H$ with capacity $N$, mini batch size $\lvert B\rvert$\;
\textbf{Initialize:} Upper limit of Radio Failures $TH_1$, UAV flight speed ;
 \For{episode = 1,...,Max episode}{
 Initialize a sliding window queue W with capacity $N_1$\;
 Initialize $q_0 = \{ \mathbf{q}_I\} \in \mathcal{S}\setminus\mathcal{S_O}$, set step $k \xleftarrow[]{} 0$\;
  \For{each step of episode}{
        Select action $a_n$ according to parameterized network $\hat{\pi}(\cdot \mid q_n; \theta_{\pi})$ \;
    Agent execute action $a_n$, observe $\{\mathbf{q_{n+1}}\}$ and $c_n$,$d_n$\;
    Store experience $(\mathbf{q}_n, a_n, c_n, d_n, \mathbf{q}_{n+1}, \delta_c, \delta_d)$ in sliding window queue W\;
    When reached a number $N_1$ of transitions, store them in replay memory $H$ and compute \eqref{eq:C_Nstep} and \eqref{eq:D_Nstep}\;
    From buffer $H$ sample minibatch $B$ of $N_1$ experience according to the priority as for \eqref{eq:priority}\; 
    Update the \ac{DNN} of state action cost function $Q_C$ performing gradient descent on loss \eqref{eq:LossQ_C} with respect to $\theta_C$\;
    Update the \ac{DNN} of state action constraint function $Q_D$ performing gradient descent on loss \eqref{eq:LossQ_D} with respect to $\theta_D$\;
    Update the priority weights $ p_{c,n}$, $p_{d,n}$ based on TD error\;
    Obtain $\pi^* $ by \eqref{eq:LP_Lyap}\;
    Update $\hat{\pi}(\cdot \mid q_n; \theta_{\pi})$ via $\theta_{\pi} \xleftarrow{} \theta_{\pi}-\alpha \nabla_{\theta_{\pi}}L(\theta_{\pi}) $\;
    }
 Update the target networks after I iterations.\\
 Set $\pi_T = \pi^*$ and $\theta_T = \theta_{\pi}$\;
 }
\end{algorithm}

\section{Transfer Learning via Teacher Policy}\label{TransferAdivce}
In this section we describe the Teacher Advice algorithm to provide external knowledge and allow the agent pre-trained in $D_1$ to quickly adapt to the new environment $D_2$.

Let us assume there exists a policy $\pi_T$, solution function of \eqref{eq:CMDP_opt_problem} mapping states to actions, in a defined domain $D_1 = \langle \mathcal{S}, \mathcal{A}, \mathcal{P}, \mathcal{C}_1, \mathcal{D}_1, \alpha, d_{th} \rangle$ at $f_1$ to get from a particular starting point to a goal, given a set of outage states. 
Let us now consider a domain $D_2$ at frequency $f_2$ that differs from domain $D_1$ by the constraint cost distribution: $ \mathcal{D}_1, \neq \mathcal{D}_2$.
We propose to consider $D_1$ as the old domain and $D_2$ as the new domain.

To reduce the computational burden of the training process in the new domain, we propose leveraging \acl{TL} to learn an optimal policy by leveraging exterior information from $D_1$ as well as internal information from $D_2$. 
\begin{figure}[t]
    \centering
    \includegraphics[]{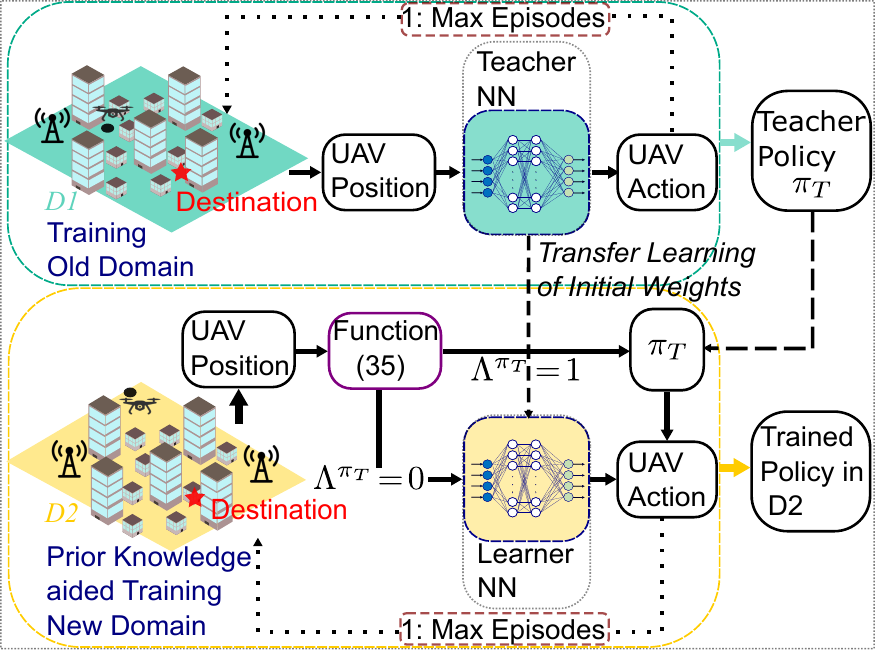} 
    \caption{Illustration of our proposed Transfer Learning Algorithm: a pre-trained policy in domain $D_1$ is used as teacher in domain $D_2$.}
    \label{fig:TransferLearningprocess}
\end{figure}
The robust teacher policy $\pi_T$ supports the exploration process in domain $D_2$ at frequency $f_2$ in two ways (\figref{fig:TransferLearningprocess}).
In the first step, we use robust trajectories generated using a pre-trained $\pi_T$ to provide prior knowledge about the task. 
To reach this goal, we utilize the concepts of known and unknown spaces \cite{garcia2012safe} to cover some regions of the feature space. 
In a second step, the agent transfers the pre-trained weights from $D_1$ and starts its training in the new domain $D_2$. The teacher policy $\pi_T$ is used to support the exploration process when the agent meets an unknown state. In new situations, the learning agent evaluates a state from the perspective of the old domain to reduce the frequency of risky states.
Here it is important to note that the role of the teacher policy is not to supply the best action but to advice an action more robust than the one obtained through random exploration.
\figref{fig:TransferLearningprocess} summarizes the overall \ac{TL} process adopted in this paper.
Note that the proposed \ac{TL} method is applicable to any \ac{DRL} algorithm and it is not specific to the robust-DDQN only.

\subsection{Initial Known Space}
The agent, equipped with an empty memory $C$ of size $Z$, builds the initially known space by storing new experiences.
Using $\pi_T$, we run $Q < Z$ iterations with the environment and collect states, actions taken, reward received (cost and constraint cost in this case), and if the current state is terminal. The stored data follows the structure of the experience replay memory used in conventional \ac{DQN}. Each memory element represents a transition the agent has experienced in domain $D_1$.
The resulting data forms the known space.
When the agent enters a new state $\mathbf{q}_n$, it computes the euclidean distance to determine if $\mathbf{q}_n$ belongs to the known space.
Hence, we define a density threshold $\Theta$ and a risk function as \cite{garcia2012safe}
\begin{equation}\label{eq:riskFunction}
\Lambda^{\pi_T}(a_n|\mathbf{\mathbf{q}_q}) =
    \begin{cases}
    0, & \mbox{if} \displaystyle\min_{1\leq q \leq Z} d_{n,q} \leq \Theta\\
    1, & \mbox{otherwise}.
\end{cases}
\end{equation}
where $d_{n,q} = \norm{\mathbf{q_{n}}-\mathbf{q_{q}}}$ is the Euclidean distance between a new state and the states in memory.
The parameter $\Theta$ defines the classification region for a new state $\mathbf{q}_n$ and it is dependent on the size of the action. In this work, we consider $\Theta = 2 \Delta_T V\textsubscript{max}$.
When the distance of the nearest neighbor to $\mathbf{q}_n$ is greater than $\Theta$, the experience is added to the memory.
 
Thus, the definition of a known state is as following:
\begin{definition}\label{Known_Unknown}
Given a density threshold $\Theta$, a state $\mathbf{q}_n$ is considered known when $\Lambda^{\pi_T}(\mathbf{q}_n) = 0$ and unknown in all other cases. 
Formally, $\Sigma \subseteq \mathcal{S}$ is the set of known states, while $\Upsilon \subseteq \mathcal{S}$ is the set of unknown states with $\Sigma  \cap \Upsilon = 0$.
\end{definition}

Using the known space set, we could transfer the learner the advice to prefer some actions over others in specific regions of the feature space. However, a direct translation of the action in the new domain would heavily limit the agent ability in domain $D_2$. 
To make our approach robust to imperfections in the advice or teacher policy, we are interested in providing the learning agent with the possibility to refine the transferred knowledge based on its subsequent trajectories in domain $D_2$.
In what follows, we present the algorithm for the training of the learner agent in domain $D_2$.
\subsection{Training in New Domain}
The algorithm to train the learning agent in the new domain is composed of an initialization step and a reinforcement learning step.
The different steps that can be summarized as follows:
\begin{enumerate}[a.]
\item \textbf{Initialization Step}:  
In this step the hyperparameters, the density threshold $\Theta$ and the initial state $\mathbf{q}_I$ are initialized.
The algorithm transfers the weights of the teacher \ac{DNN} pre-trained in $D_1$ to domain $D_2$, in \ac{DNN} networks with identical structure. In addition, to obtain new and improved ways to complete the task, we add Gaussian noise to the initial weights such that $\theta_{\pi_{f_2}} = \theta_{\pi_T} + \mathcal{N}(0,\sigma^2)$. 
\item \textbf{Reinforcement Learning Step}: In this step, the training in $D_2$ starts and the algorithm refines the policy to satisfy the connectivity constraint in domain $D_2$.
When the UAV flies in a new position, if the state is known, $\mathbf{q}_n \in \Sigma$, the agent performs an action $a_n$ using policy network  $\hat{\pi}(\cdot \mid q_n; \theta_{\pi_{f_2}})$ and train the networks in domain $D_2$. 
In unknown states, instead, the action $a_n$ is performed using the teacher policy $\pi_T$ and the experience is added to the known set in memory $Z$. As the exploration process and the training in $D_2$ continue, the knowledge of the agent of $D_2$ and the accuracy of $\pi_2$ improve. 
Hence, the algorithm utilizes the teacher policy $\pi_T$ only as a backup policy with to guide the learning away from risky states or, at least, reduce their frequency.
\end{enumerate}

The pseudo code for the Transfer Learning and Teacher advice is reported in Algorithm \ref{alg:TrasferLearningDQN}.

\begin{algorithm}[t]
\SetAlgoLined
Given baseline behavior $\pi_T$ and memory $C$ with maximum size $Z$\;
\textbf{Initialize} maximum number of episodes, density threshold $\Theta$, prioritized replay memory $H$ with capacity $N$,  mini batch size $\lvert B\rvert$\;
Create $\Sigma$ collecting $Q$ interactions\; 
Transfer Initial weights from $\pi_T$\;
\textbf{Set}  maxTotalRwEpisode = 0\;
 \For{episode = 1,...,Max episode}{
 Initialize $q_0 = \{ \mathbf{q}_I\} \notin \mathcal{S_O}$, set step $k \xleftarrow[]{} 0$\;
  \For{each step of episode}{
  Compute the closest $\mathbf{q}_q \in C$ to $\mathbf{q}_n$ using \eqref{eq:riskFunction}\;
  \If{($\mathbf{q}_n$ is known)}{
        Select action  using $\pi = \hat{\pi}(\cdot \mid q_n; \theta_{\pi_{f_2}})$\;
        Execute lines 8-15 of Algorithm \ref{alg:SafeDDQN}\;
  \Else{
         Choose an action using $\pi_T$\;
         Agent execute action $a_n$, observe $\{\mathbf{q_{n+1}}\}$ and $c_n$,$d_n$\;
         Add experience to memory $C$;}}
    
    Remove least frequently used experiences in $C$\;
    }
     Update the target networks after I iterations.\\
    Set $\pi_{2} = \pi^*$\;
 }
 \caption{Transfer Learning for Connectivity-Aware Path Design}\label{alg:TrasferLearningDQN}
\end{algorithm}


\section{Numerical Results}\label{NumericalResults}
In this section we present the main numerical results of our findings. 
We first describe the radio environment used for generating the \ac{UAV} trajectories. Then, we evaluate the performance of the proposed robust-\ac{DDQN} algorithm in domain $D_1$ at $f_1$.
We compare our approach with state of the art deep \ac{RL}. 
Specifically, we implement an unconstrained Dueling \ac{DDQN} that has been shown to suit \ac{UAV} connectivity-aware path problems \cite{zeng2020simultaneous, fontanesi2021deep}. We model the reward function to minimize the flight time and the number of radio failures for a fair comparison.
Details about the implementation of the Dueling \ac{DDQN} benchmark strategy will be presented in Appendix \ref{DDQN_Implementation}. 
At last, we validate and show the benefit of the transfer learning approach from $D_1$ at $f_1$ to  $D_2$ at $f_2$.
\subsection{Radio Environment}
The radio environment where the \ac{UAV} is flying is composed of buildings generated based on the International Telecommunication Union (ITU) model \cite{ITUBlockageModel}, which involves three parameters: i)the ratio of land area covered by buildings to total land area, ii) the mean number of buildings per unit area, iii) the height of buildings modeled by a Rayleigh \ac{PDF}. 
The above parameters can be modified as specified in \cite{holis2008elevation} to create Suburban, Urban, Dense Urban and High Rise Urban environments. We have considered the last three mentioned environments as they are the most challenging for connectivity-aware UAV path and to demonstrate the generality of our approach.
Each environment has a different \ac{BS} number, \ac{BS} power and height within a geographical area of $L \times L$, as for \ac{BS} density specified in \cite{3GPPStudy_macroBS}. 
At frequency $f_1 = 2$ GHz, we consider 8 antenna elements at the ground \ac{BS}, while 64 antennas at $f_2 = 28$ GHz \cite{rebato2018study}, as for the \ac{ULA} and \ac{UPA} antenna models described in Section \ref{AntennaModel}.
At sub-6 GHz, we adopt the \ac{3GPP} Macro Path Loss Model for Urban scenario \cite{3GPPStudy_LTE_aerial}, that includes modeling for \ac{LoS} and \ac{NLoS} channels.
The presence/absence of obstacles is determined in the simulated environment by checking whether the line BS-UAV is blocked or not by any building. A ray tracing software would allow us to include in the propagation calculation the relative permittivity and conductivity of the surface material, which is different for any building. However, this information would limit the algorithm's training to a specific scenario or condition. The statistical ITU building model \cite{ITUBlockageModel} used in our approach reflects the average characteristics over a large number of geographic areas of similar type and has been widely used to characterize urban environments in UAV trajectory path design \cite{zeng2020simultaneous,esrafilian2019_trajectory_3dMap_SINR}.
Using data extracted by a simulator allows us to train the proposed robust-DDQN and transfer learning method on a broader general scenario, improving the algorithm's generalisation.
At \ac{mmWave} we consider the path loss model in \eqref{eq:pathLoss_f2} with $\alpha_L = 2$, $\alpha_{NL} = 4$, $X_L$, $X_{NL} = 5e^{-4}$.
We have adopted a bandwidth of 10 MHz at sub-6 GHz, $100$ MHz at \ac{mmWave} and a transmit power of 36 dBm at sub-6 GHz and 30 dBm at \ac{mmWave}, which are in line with the specifications envisioned for downlink transmission in \ac{5G} \ac{mmWave} mobile networks.
We consider a \ac{UAV} speed of $20$ m/s \cite{chen2021joint_Augmented} and ease of illustration but without loss of generality, a fixed fly altitude.

The remaining simulation parameters can be found in Table \ref{table:SimulationParameter}.
\begin{table}[t]
\caption{Parameters utilized in the simulation environment}
  \centering
\begin{tabular}{|p{1.15cm}|p{3.55cm}|p{2.05cm}|}
 \hline
 \multicolumn{3}{|c|}{Radio Simulation Parameters} \\
 \hline
 Parameter & Description & Value\\
 \hline
  $L$ & Area Size & 1 [km] \\
  $V\textsubscript{max}$ & UAV Speed  & 20 [m/s] \\
  $ h_{n}$ & UAV Height & 100 [m] \\
    $\phi_1$/ $\phi_2$ & Antenna Tilt $f_1$/$f_2$ & -10/10$\degree$ \\
    $G_{max}$ & max directional gain 
antenna element  & 8 dBi \\
  $\sigma^2$ &  Noise Power sub-6/mmWave &  -204/-120 [db/Hz] \\
  $m_v$ &  Nakagami Fading param. &  3 \\
  $\Delta_T$ & Time Step Length & 0.5 [s]\\
  $\bar{\gamma}$ & SINR Threshold & 0 dB\\
  $\bar{P}_{th}$ & Ouatge Threshold & 0.9\\
  $d_{th}$ & Connectivity Outage Threshold & 10$\%$\\
 \hline
\end{tabular}
\label{table:SimulationParameter}
\end{table}

\begin{table}[t]
\caption{Hyperparameters utilized in the simulation environment}
  \centering
\begin{tabular}{|p{1.15cm}|p{3.6cm}|p{2cm}|}
 \hline
 \multicolumn{3}{|c|}{Hyperparameters robust-DDQN} \\
 \hline
 Parameter & Description & Value\\
 \hline
  $N_1$  & n-STEP & 10\\
  $\overline{N}$ & Max Steps & 100\\
  $\alpha$ &  Discount Factor & 0.99 \\
  $\vert B \vert$ &  Minibatch Size & 32 \\
  $H$ & Replay Memory Size & 200000\\
  c & Move Penalty & 0.5\\
  d & Radio Failure Penalty & 1\\
  $\eta$ & Priority Sample Weight & 0.9 \cite{kapturowski2018recurrent}\\
 \hline
\end{tabular}
\label{table:HyperParameters}
\end{table}

\subsection{Performance of the robust Teacher Policy}
\begin{figure}[t]
    \centering
    \includegraphics[width=1\columnwidth]{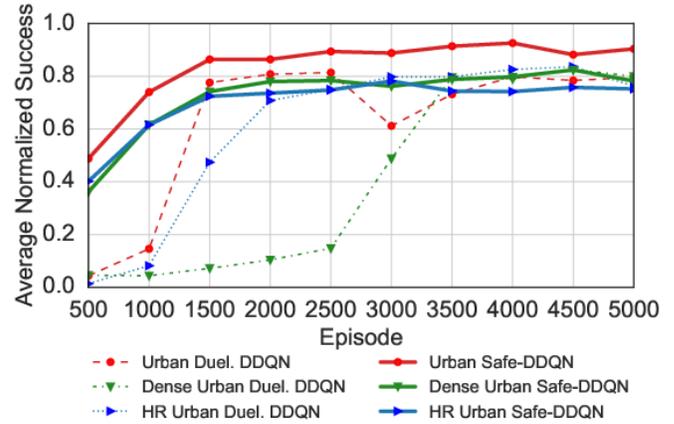}
    \caption{Results of the robust-\ac{DDQN} compared with Dueling DDQN for three urban environment with different building distribution. Term HR stands for High Rise}
    \label{fig:SafeDQN_vs_DDQN}
\end{figure}

\begin{figure}
     \centering
     \begin{subfigure}[b]{4.35cm}
         \centering
         \includegraphics[width=4.35cm]{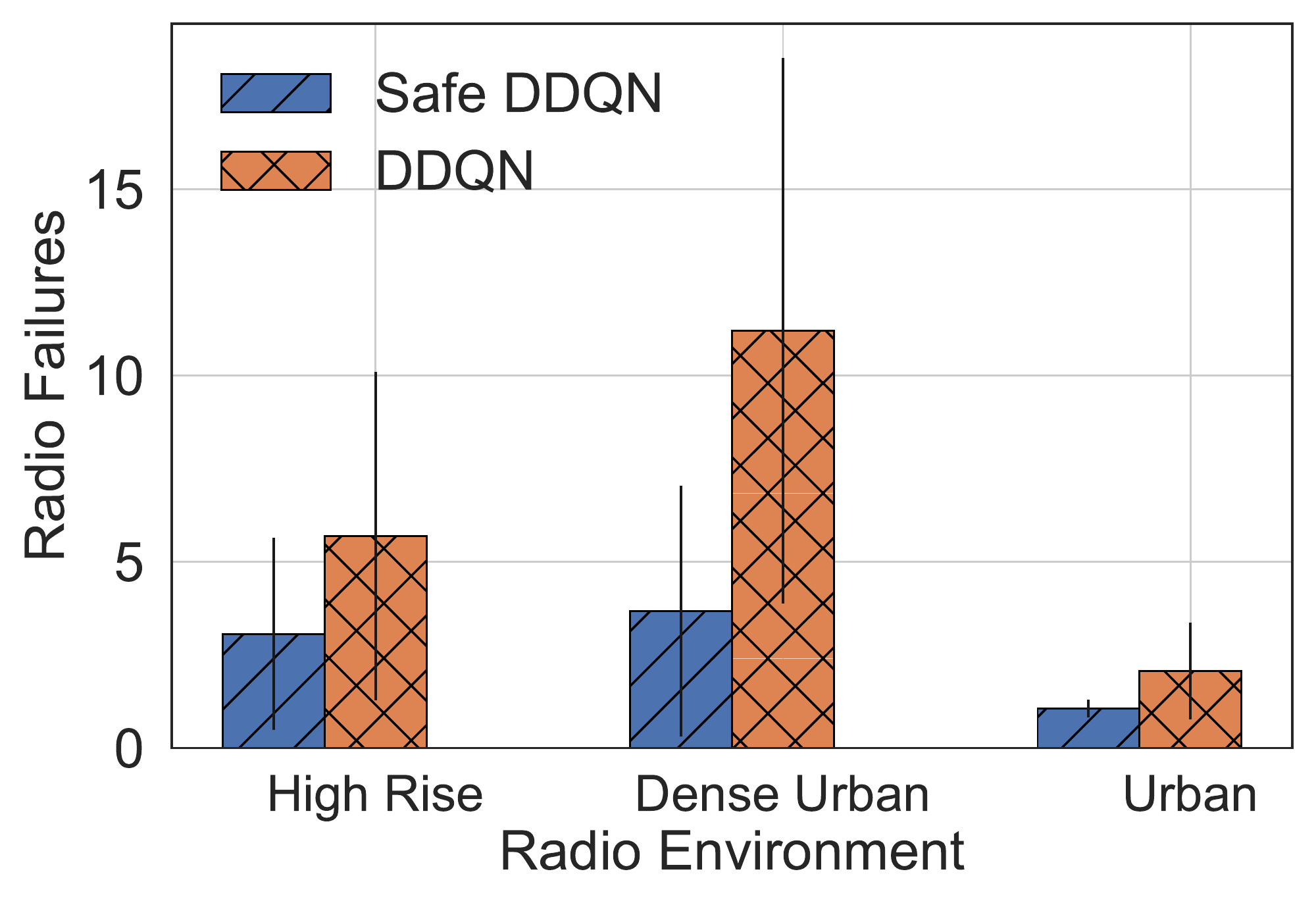}
    \caption{UAV navigation constraint cost satisfaction comparison}
    \label{fig:SafeDQN_vs_DDQN_barFailures}
     \end{subfigure}
     \hfill
     \begin{subfigure}[b]{4.35cm}
         \centering
         \includegraphics[width=4.35cm]{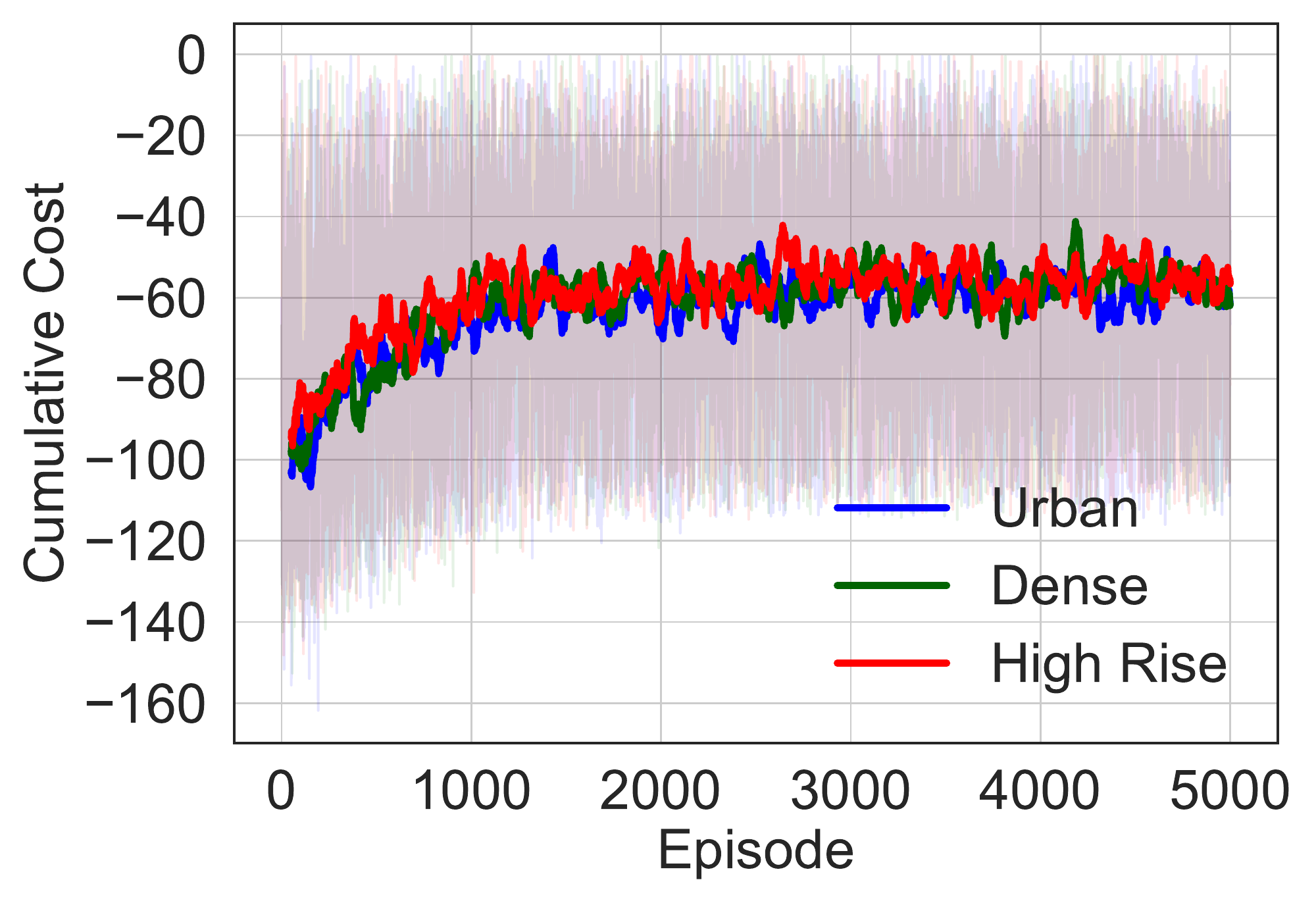}
    \caption{Average cumulative reward for three different urban environments}
    \label{fig:CumulativeReward_withSTD}
     \end{subfigure}
        \caption{Convergence of the proposed robust DDQN algorithm: (a) maximum radio failure satisfaction  and (b) reward control}
        \label{fig:Reward_RadioFailures}
\end{figure}

In this section, we show the performance of the robust Teacher Policy derived using a robust-\ac{DDQN} approach.
Table \ref{table:HyperParameters} shows some hyperparameters used to generate the results.
More details about the implementation of the robust-\ac{DDQN} approach can be found in Appendix \ref{SafeDQN_Implementation}.
The mission is considered successful if the \ac{UAV} reaches the destination before the constraint threshold is exhausted.
The destination is placed in $\mathbf{q}_F = [700,800,100]$ m and the actions at each step are left-right-forward-back.
To make the path task more challenging, we consider in \figref{fig:SafeDQN_vs_DDQN} a conservative constraint threshold $d_{th}=10$.
Note that the training phase of the robust-DDQN model is executed for a number of 5000 episodes, each of which accounts for a maximum of $N=100$ steps.
The fairness of the experiment episodes is ensured by running the trials with a different preset random seed.
The mission success rate is averaged over 500 evaluation episodes with a random initial starting point. The initial starting point is chosen from a continuous space in the area $L \times L$. Note that, to ease the algorithm generalization, the initial position is not fixed but is chosen randomly in the flying area for each evaluation episode. We also mention that the final position is chosen inside a fixed area of side $\pm \Delta = 30$ m.
\figref{fig:SafeDQN_vs_DDQN} shows the normalized success rate for the proposed robust-\ac{DDQN} compared with a conventional unconstrained Dueling \ac{DDQN}. The x-axis shows the number of episodes, while the y-axis shows the mission success.
While both algorithms converge with good performance, the proposed robust-\ac{DDQN} algorithm has a generally higher success rate. In addition \figref{fig:SafeDQN_vs_DDQN_barFailures} shows that our Lyapunov-based algorithm can control the radio failures even when the environment is more challenging. On the contrary, the unconstrained benchmark \ac{DDQN} is more apt to violate the constraint during training.

\begin{figure}[ht]
    \centering
    \includegraphics[width=1\columnwidth]{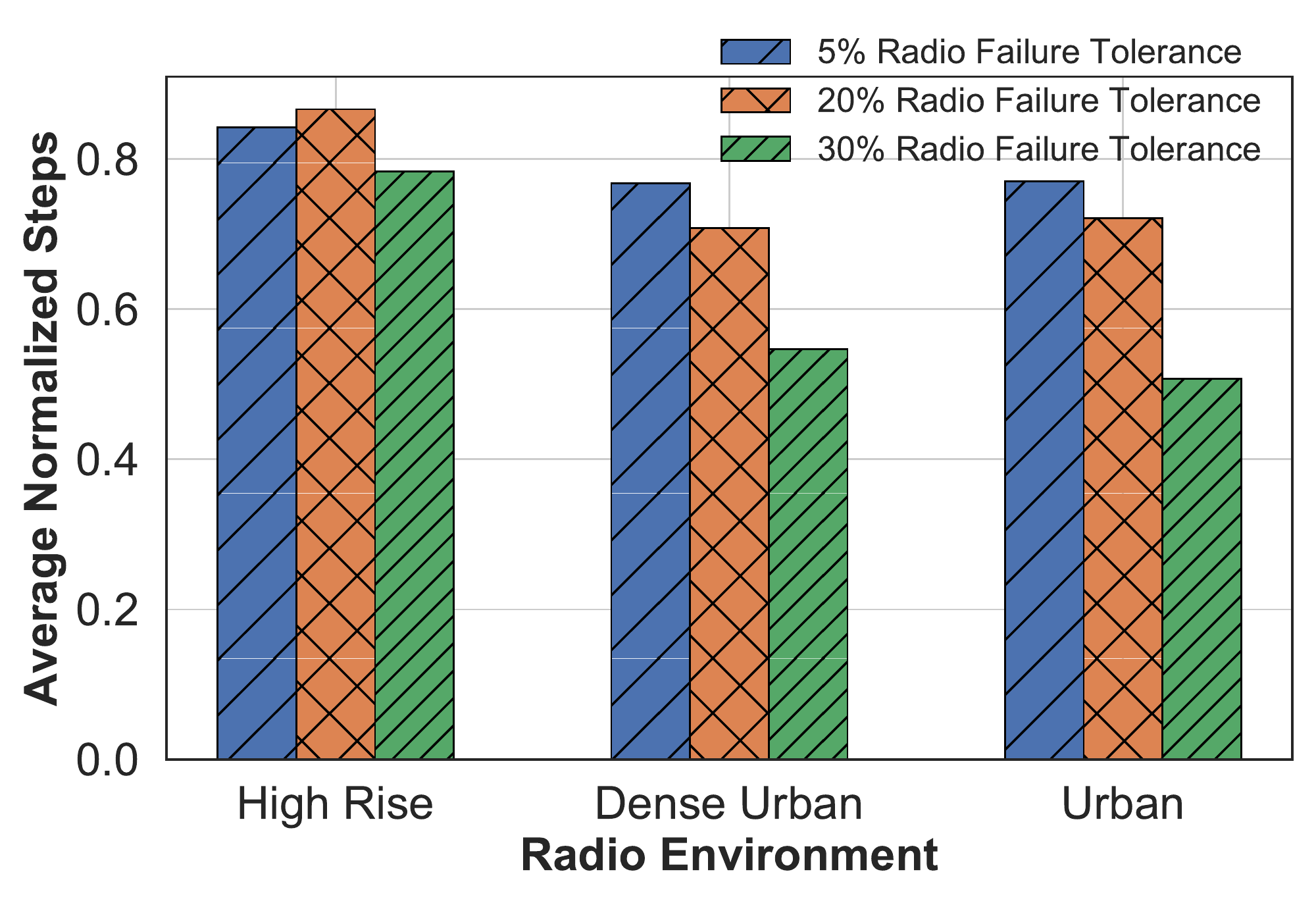}
    \caption{Impact of the Connectivity Outage Threshold on the proposed robust DDQN path Design}
    \label{fig:ConnectivityOutageImpact}
\end{figure}

\figref{fig:CumulativeReward_withSTD} shows the reward received by the agent for different urban environments. The robust-\ac{DDQN} can adequately learn the path design task with good return while satisfying the connectivity requirement.
The shaded areas in \figref{fig:CumulativeReward_withSTD} represent the 1-SD confidence intervals over 500 runs.  Finally, \figref{fig:ConnectivityOutageImpact} evaluates how the method generalizes to different values of connectivity outage threshold $d_{th}$. 
Conservative thresholds lead to longer trajectories, while higher $d_{th}$ allow more flexibility and shorter trajectories. 

\subsection{Performance of Transfer Learning}
\begin{figure}[t]
    \centering
    \includegraphics[width=1\columnwidth]{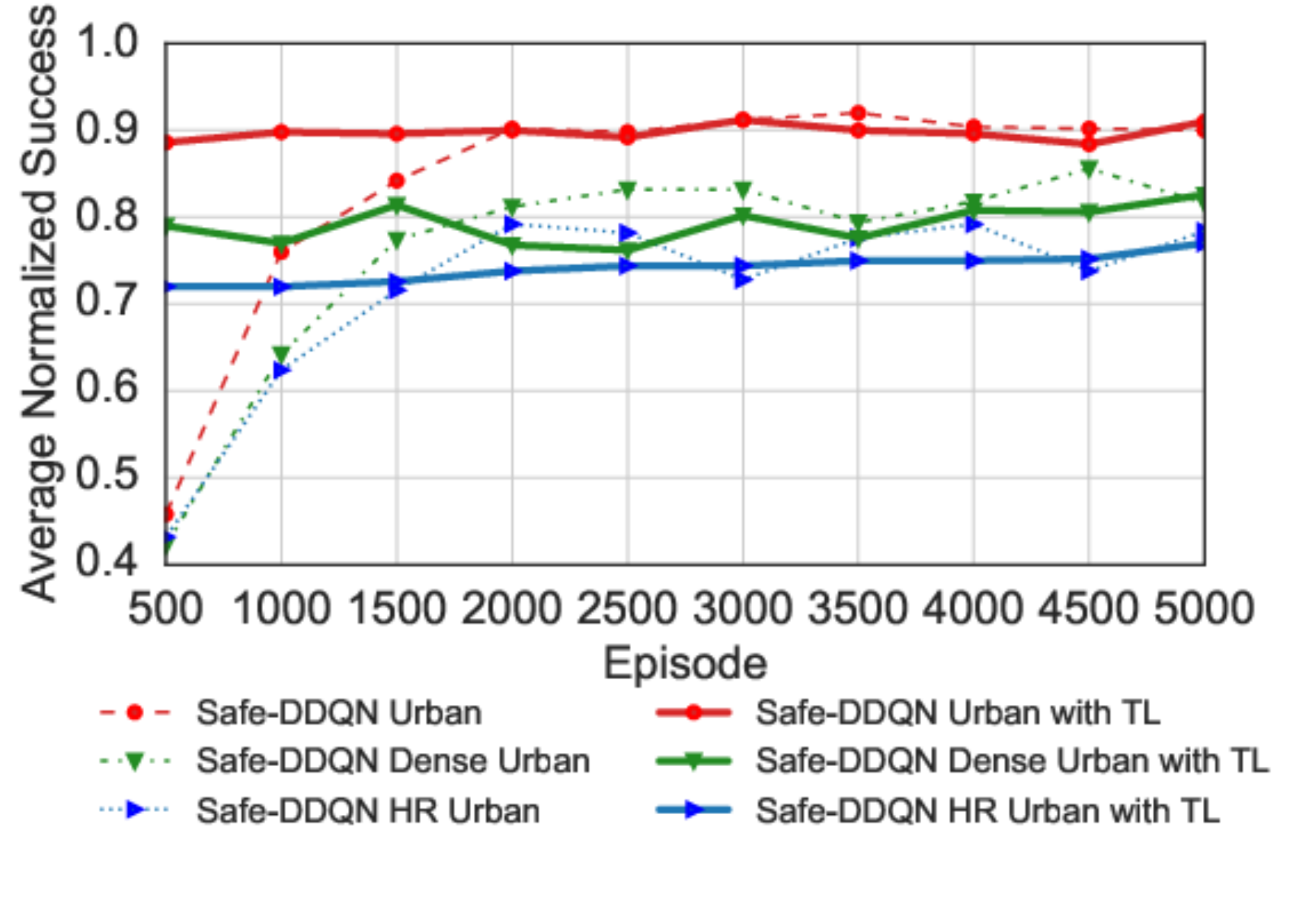}
    \caption{robust-DDQN: Average Asymptotic Performance of the Transfer Learning algorithm measured in $\%$ of accomplished missions for different urban environments at mmWave.}
    \label{fig:TL_accuracy_SafeDQN}
\end{figure}

In this subsection we investigate the potential of the transfer learning algorithm in domain $D_2$. Details about the implementation of the Teacher advice and Transfer Learning algorithm are in Appendix \ref{TL_Implementation}.
The impact of the transfer learning is measured considering the asymptotic performance of the agent at \ac{mmWave}. The TL algorithm is executed for 5000 trials and is compared with the algorithm executed without TL.
The mission rate is again averaged over 500 episodes with the same constraint threshold as the previous section.
We investigate first the case of using a teacher policy pre-trained in sub-6 GHz via robust-\ac{DDQN}.
The results are shown in \figref{fig:TL_accuracy_SafeDQN}.
The curves show the average mission success of over 500 episodes.
The transfer learning is here very effective since the algorithm with \ac{TL} needs few training trials to reach the asymptotic performance of the algorithm trained tabula rasa.

\begin{figure}[ht]
    \centering
    \includegraphics[width=1\columnwidth]{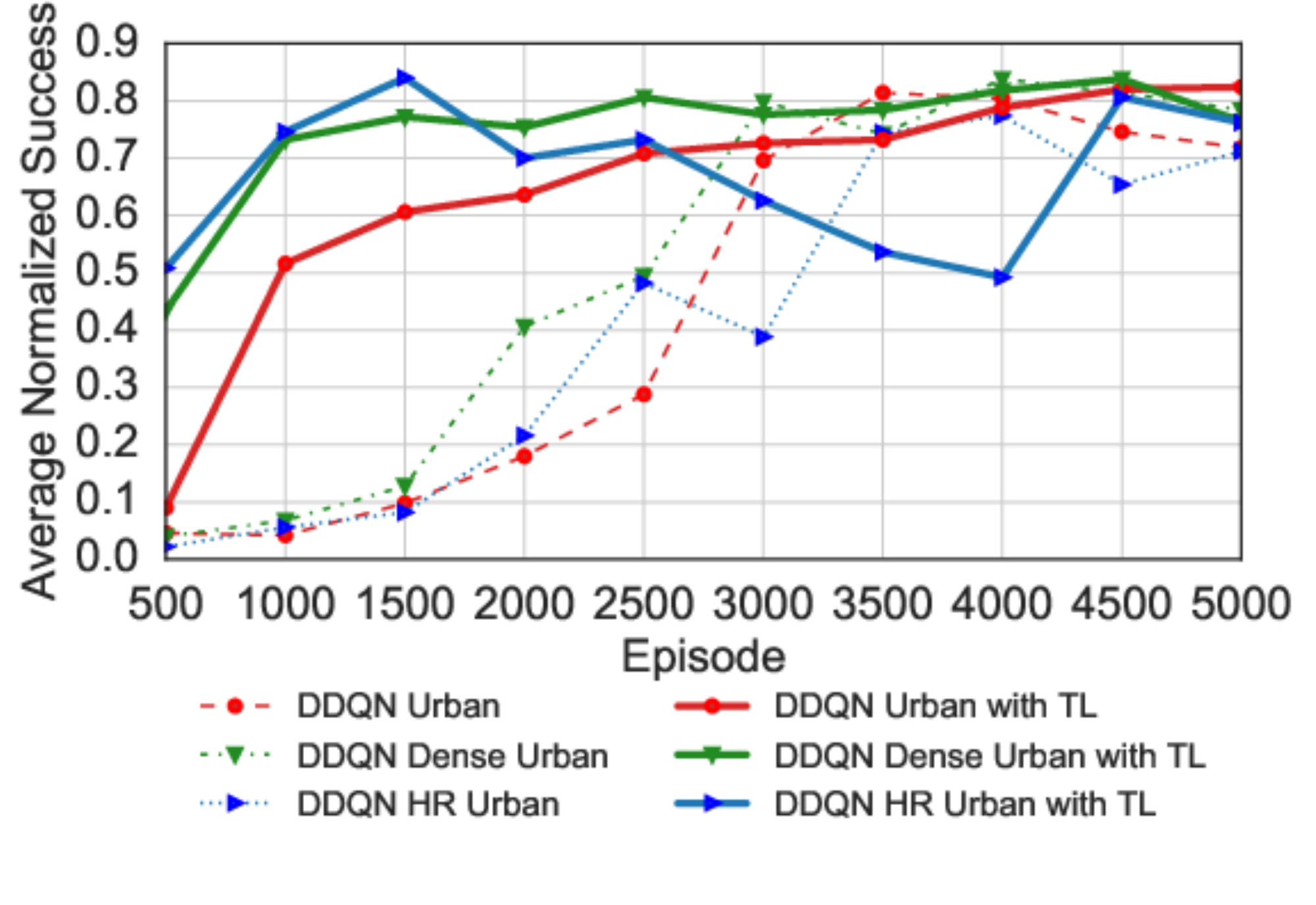}
    \caption{Dueling DDQN: Average Asymptotic Performance of the Transfer Learning algorithm measured in $\%$ of accomplished missions for different urban environments at mmWave}
    \label{fig:TL_accuracy_DDQN}
\end{figure}

In addition, \figref{fig:TL_accuracy_DDQN} shows the results of the teacher advice transfer approach using a Dueling \ac{DDQN} as a teacher.
Transfer Learning is again very powerful, as the Dueling DDQN without transfer needs at least 300 episodes to perform comparably to the algorithm with transfer.

\begin{figure}
     \centering
     \begin{subfigure}[b]{6cm}
         \centering
         \includegraphics[width=6cm]{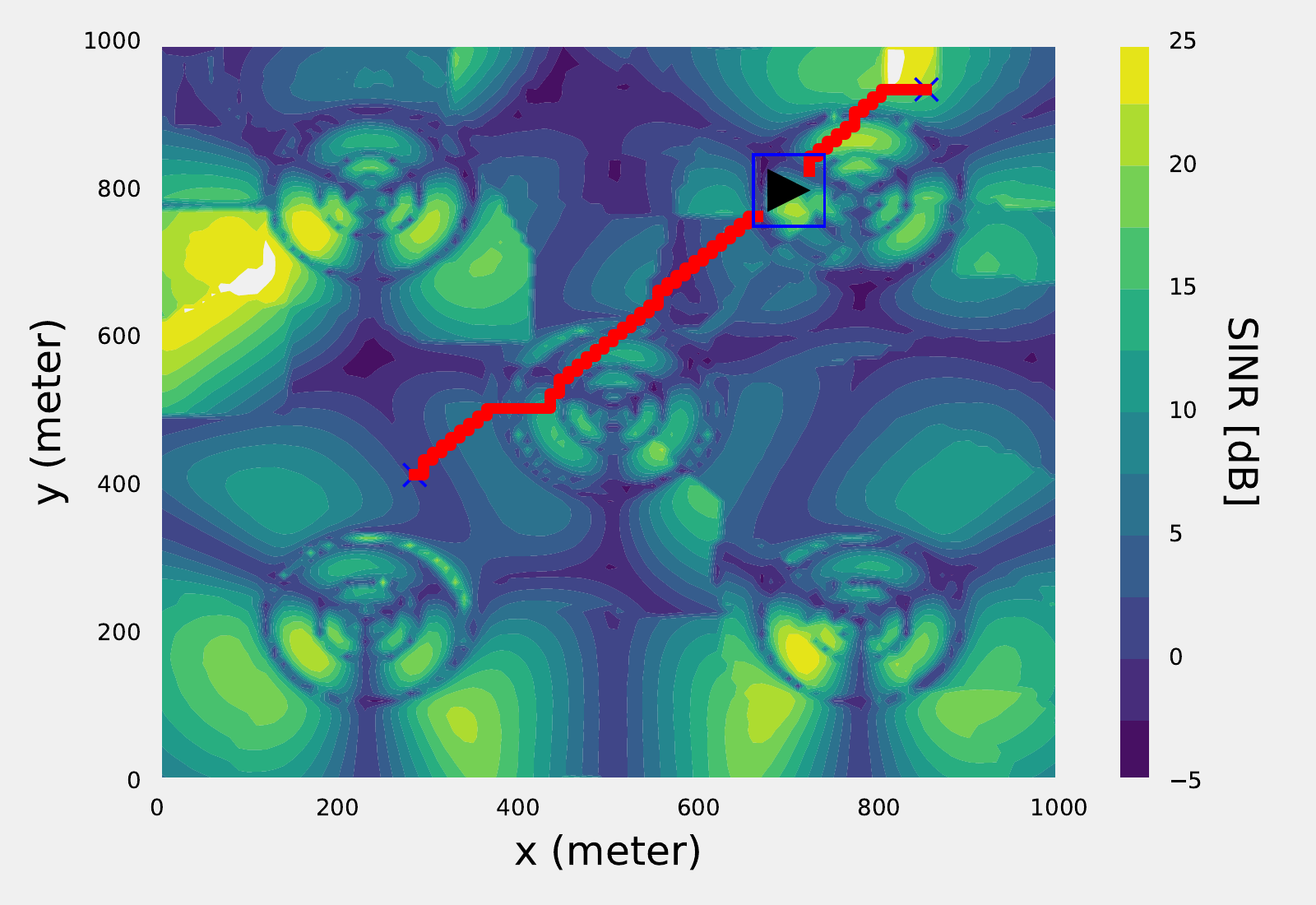}
    \caption{UAV path design example at sub-6 GHz along with the radio map. The UAV reaches the destination from two different starting points during the training in the old domain.}
    \label{fig:radioMapsub6}
     \end{subfigure}
     \begin{subfigure}[b]{6cm}
         \centering
         \includegraphics[width=6cm]{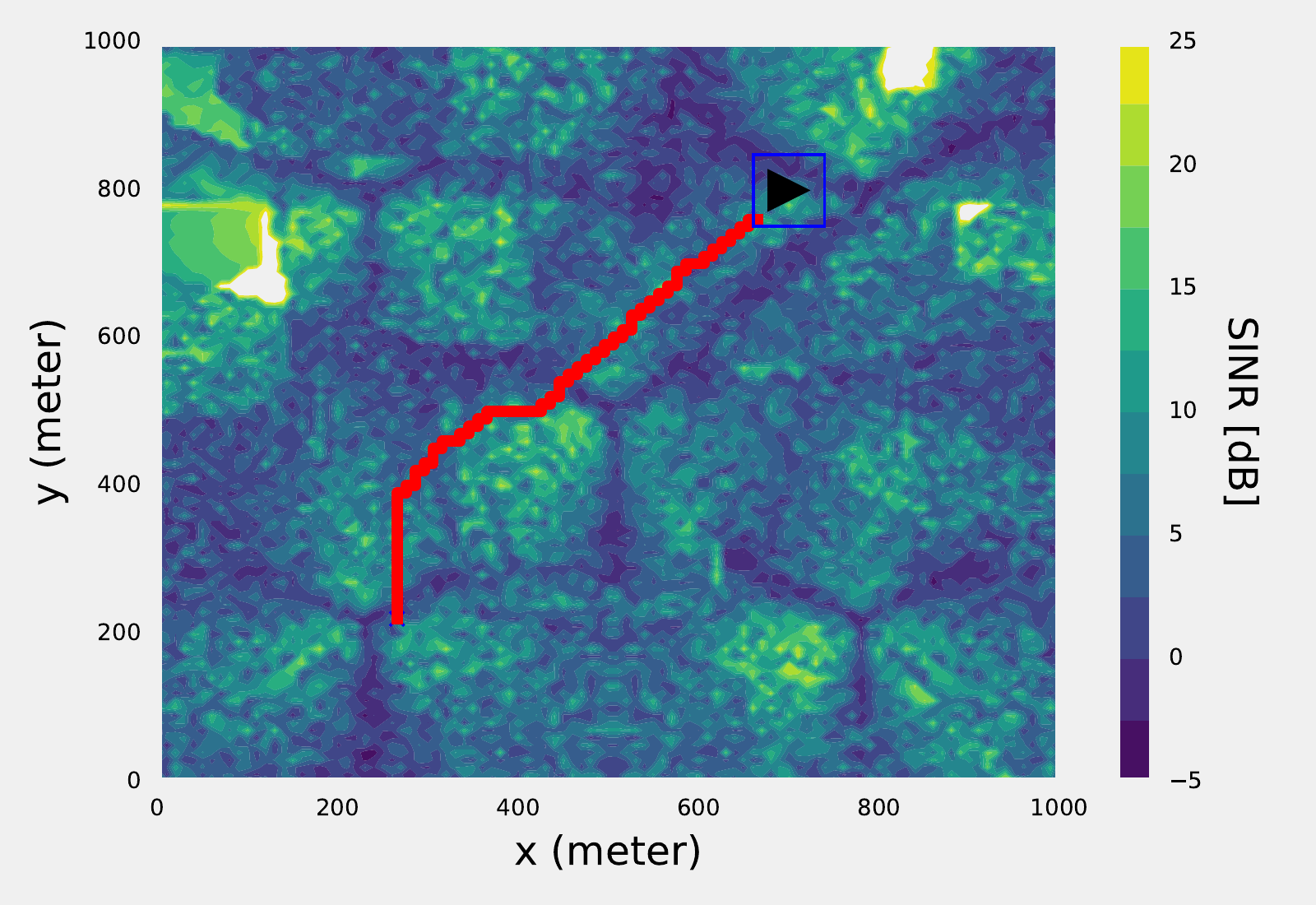}
    \caption{Returned UAV path design in the new domain at mmWave along with the radio map.}
    \label{fig:radioMapmmWave}
     \end{subfigure}
        \caption{An illustrative example of the radio maps in a High Rise environment for the sub-6 GHz and mmWave and the returned UAV path. Radio failures occur below the SINR threshold of 0 dB.}
        \label{fig:RadioMaps_paths}
\end{figure}

\figref{fig:RadioMaps_paths} shows an example of the radio map for the High Rise environment. \figref{fig:RadioMaps_paths} is coloured according to the average \ac{SINR}. Lighter colour means a higher SINR and vice versa. 
Generally, it is visible a different behaviour between the two bands. At Sub-6GHz, lower SINR is in interference regions between the \acp{BS}. At mmWave, the lower SINR areas are more irregular due to the combined effects of the higher BS antenna tilt and building blockage.
In \figref{fig:radioMapsub6} we plot the radio map at sub-6 GHz for a UAV height of 100 m together with two paths that start from two different initial points. The UAV reaches the destination from two different starting points during the training in the old domain.
Recalling that a radio failure occurs at average SINR values below the SINR threshold $0$ dB, it is possible to see that the \ac{UAV} adjusts its trajectory to satisfy the connectivity constraint.
In \figref{fig:radioMapmmWave} we plot the radio at mmWave band for the same UAV height and the returned path. The UAV is reusing some of the previous knowledge to reach the destination.  

In conclusion, results show that the transfer advice framework proposed in this paper helps a learner agent reduce the training time in successful missions using both the proposed robust-\ac{DDQN} and a conventional Dueling \ac{DDQN} as a teacher. Different environments with different levels of complexity in terms of coverage aware UAV navigation have been tested.
This shows that the proposed \ac{TL} framework is versatile and not dependent on the algorithm used to train the teacher policy. However, it is important to observe that the robust-DDQN, creating a policy that respects the connectivity constraint throughout training, results in a better teacher policy.

\section{Conclusion}\label{Conclusion}
In this paper, we have developed a \ac{DDQN} Lyapunov based approach to solve the non-convex \ac{UAV} connectivity-aware path design across different simulation environments. We then proposed a \acl{TL} technique to improve the agent learning in a new domain at \ac{mmWave} using the knowledge gained in a domain $D_1$ at sub-6 GHz.
We have evaluated the efficiency of our \ac{TL} approach using a Lyapunov based DDQN teacher policies derived at sub-6 GHz benchmarked with a Dueling \ac{DDQN}.
Our approach showed the potential of the proposed \ac{TL} framework to save many training episodes for both the teacher policies, resulting in fewer \ac{UAV} flights. 
The learning agent's convergence using a teacher policy derived via the Lyapunov based \ac{DDQN} is faster for all the different urban scenarios under consideration.
Future works include the evaluation of the sensitivity of our algorithm to the advice of a non-perfectly trained teacher.

\section*{Acknowledgement}
The authors wish to thank Mr. Brian Keogh from University College Dublin for the helpful discussions. The authors also thank the Editor and anonymous reviewers for their constructive comments.

\appendices

\section{Robust DDQN Implementation}\label{SafeDQN_Implementation}
Table \ref{table:NeuralNetworkParameters} displays the architecture of the neural networks used in our robust-DDQN algorithm. Especially, Table \ref{table:NeuralNetworkParameters} shows the learning rate values adjusted in the algorithm to reach convergence.
The hyperparameters in Table \ref{table:HyperParameters} are selected to both achieve a good trade off between learning performance and model complexity.
We implement the proposed robust-DDQN based on Tensorflow library in Python.
The Cost and Constraint Cost layers are all fully connected and consist of four hidden layers with ReLU as an activation function. The layers have respectively 64, 64, 32 and 4 nodes, respectively.
The policy network consists of eight hidden layers, activated with ReLu and with respectively 512, 256, 128, 128, 64, 64, 32 nodes.  
Weights of the policy network are initialized using the inverse distance from the \ac{UAV} location to the destination, that is  considered known, so that $\hat{\pi}(\cdot \mid q_n; \theta_{\pi})$ approximates $\frac{1}{\norm{q'-\mathbf{q}_F}}$, where $q'$ is the next state after taking action $a_n$. The policy networks weights are updated each 5 episodes, while the Cost and Constraint Cost's ones each 25 episodes.
Adam optimizer \cite{diederik2014adam} is used to apply gradient descent for all the networks. The learning rate is reported in Table \ref{table:NeuralNetworkParameters}. 

\begin{table}[t]
\caption{Networks Structure in robust-DDQN Algorithm}
  \centering
\begin{tabular}{|p{1.5cm}|p{1.8cm}|p{1.8cm}|p{1.8cm}|}
 \hline
 Type & Output Size & Activation & Learning Rate\\
 \hline
 Reward  & $\dim(\mathcal{A})$ & Linear &  $10^{-4}$\\
 Cost  &  $\dim(\mathcal{A})$ & Linear &  $10^{-4}$\\
 Policy & $\dim(\mathcal{A})$ & Softmax & $10^{-6}$\\
 \hline
\end{tabular}
\label{table:NeuralNetworkParameters}
\end{table}

\section{DDQN Implementation}\label{DDQN_Implementation}
The \ac{DNN} of the \ac{DDQN} used for benchmark consists in a Dueling architecture with input layer, four hidden layers, one output layer, all fully connected feedforward, activated using Rectified Linear Units (ReLU) and trained with Adam optimizer to minimize the MSE. The learning rate is kept 0.01. The number of neurons of the hidden layers are 512, 256, 128 and 128. The dueling architecture represents two separate estimators, one neuron for the state value function and $K$ for the action advantages for the $K$ actions. The output of the K+1 neurons represents the aggregated output layer to estimate the $K$ action values.
The replay memory and memory $C$ for the transfer learning have size 100,000. At \ac{mmWave} we encourage exploration through Gaussian noise $\mathcal{N}$(0, 0.1) to the weights of the network.

\section{Teacher Advice and Transfer Learning Algorithm Implementation}\label{TL_Implementation}
The set of known cases $\Sigma$ is created running a number $N=250$ trajectories using the teacher policy $\pi_T$ and collecting $Q = 9000$ iterations with the environment. The teacher policy might be derived either via the Lyapunov approach or the conventional \ac{DDQN} described in the previous sections. The memory $C$ has size 200,000. 
Thus, in a second phase, the network models trained in $D_1$ are translated into domain $D_2$. Here, the weights of the networks are perturbed with Gaussian noise, $\mathcal{N}$(0, 0.1).
The training is computed using a prioritized memory of same size as for \ref{NumericalResults}.

\bibliographystyle{IEEEtran}
\bibliography{References.bib}

\end{document}

%% file: acronyms.tex
\begin{acronym} 
\acro{3G}{Third Generation}
\acro{4G}{Fourth Generation}
\acro{5G}{Fifth Generation}
\acro{3GPP}{3rd Generation Partnership Project}
\acro{BB}{Base Band}
\acro{BBU}{Base Band Unit}
\acro{BER}{Bit Error Rate}
\acro{BH}{Backhaul}
\acro{BPP}{Binomial Point Process}
\acro{BS}{Base Station}
\acro{BW}{bandwidth}
\acro{C-RAN}{Cloud Radio Access Networks}
\acro{CAPEX}{Capital Expenditure}
\acro{CBR}{case-based reasoning}
\acro{CoMP}{Coordinated Multipoint}
\acro{CMDP}{Constrained Markov Decision Process}
\acro{CPRI}{Common Public Radio Interface }
\acro{CU}{Centralized Unit}
\acro{D2D}{Device-to-Device}
\acro{DAC}{Digital-to-Analog Converter}
\acro{DAS}{Distributed Antenna Systems}
\acro{DBA}{Dynamic Bandwidth Allocation}
\acro{DL}{Downlink}
\acro{DNN}{Deep Neural Network}
\acro{DQN}{Deep Q-Network}
\acro{DDQN}{Double Deep Q-Network}
\acro{DQfD}{Deep Q-Learning from Demonstration}
\acro{DRL}{Deep Reinforcement Learning}
\acro{DU}{Distributed Unit}
\acro{FBMC}{Filterbank Multicarrier}
\acro{FEC}{Forward Error Correction}
\acro{FH}{Fronthaul}
\acro{FL}{Federated Learning}
\acro{FFR}{Fractional Frequency Reuse}
\acro{FSO}{Free Space Optics}
\acro{GSM}{Global System for Mobile Communications}
\acro{HAP}{High Altitude Platform}
\acro{HetNet}{Heterogeneous Network}
\acro{HL}{Higher Layer}
\acro{HARQ}{Hybrid-Automatic Repeat Request}
\acro{KL}{Kullback-Leibler}
\acro{KPI}{Key Performance Indicator}
\acro{kg}{Kilogramm}
\acro{IoT}{Internet of Things}
\acro{LAN}{Local Area Network}
\acro{LAP}{Low Altitude Platform}
\acro{LfD}{Learning From Demonstration}
\acro{LL}{Lower Layer}
\acro{LoS}{Line of Sight}
\acro{LTE}{Long Term Evolution}
\acro{LTE-A}{Long Term Evolution Advanced}
\acro{LP}{Linear Programming}
\acro{MAC}{Medium Access Control}
\acro{MAP}{Medium Altitude Platform}
\acro{MC}{Monte Carlo}
\acro{MDP}{Markov Decision Process}
\acro{ML}{Medium Layer}
\acro{MME}{Mobility Management Entity}
\acro{mmWave}{millimeter Wave}
\acro{MIMO}{Multiple Input Multiple Output}
\acro{ML}{Machine Learning}
\acro{MSE}{Mean Square Error}
\acro{NFP}{Network Flying Platform}
\acro{NFPs}{Network Flying Platforms}
\acro{NN}{Neural Network}
\acro{NLoS}{Non-Line of Sight}
\acro{RL}{Reinforcement Learning}
\acro{NR}{New Radio}
\acro{OFDM}{Orthogonal Frequency Division Multiplexing}
\acro{PAM}{Pulse Amplitude Modulation}
\acro{PAPR}{Peak-to-Average Power Ratio}
\acro{PDF}{Probability Density Function}
\acro{PER}{Prioritized Experience Replay}
\acro{PGW}{Packet Gateway}
\acro{PHY}{physical layer}
\acro{PP}{Poisson Process}
\acro{PSO}{Particle Swarm Optimization}
\acro{PTP}{Poin to Point}
\acro{QAM}{Quadrature Amplitude Modulation}
\acro{QoE}{Quality of Experience}
\acro{QoS}{Quality of Service}
\acro{QPSK}{Quadrature Phase Shift Keying}
\acro{RF}{Radio Frequency}
\acro{RN}{Remote Node}
\acro{RAU}{Remote Access Unit}
\acro{RAN}{Radio Access Network}
\acro{RRH}{Remote Radio Head}
\acro{RRU}{Remote Radio Unit}
\acro{RRC}{Radio Resource Control}
\acro{RRU}{Remote Radio Unit}
\acro{RSS}{Received Signal Strength}
\acro{RSRP}{Reference Signals Received Power}
\acro{RU}{Remote Unit}
\acro{SCBS}{Small Cell Base Station}
\acro{SDN}{Software Defined Network}
\acro{SINR}{Signal-to-Noise-plus-Interference Ratio}
\acro{SIR}{Signal-to-Interference Ratio}
\acro{SNR}{Signal-to-Noise Ratio}
\acro{SON}{Self-organising Network}
\acro{TETRA}{Trans-European Trunked Radio}
\acro{TL}{Transfer Learning}
\acro{TD}{Temporal Difference}
\acro{TDD}{Time Division Duplex}
\acro{TD-LTE}{Time Division LTE}
\acro{TDM}{Time Division Multiplexing}
\acro{TDMA}{Time Division Multiple Access}
\acro{UE}{User Equipment}
\acro{UAV}{Unmanned Aerial Vehicle}
\acro{ULA}{Uniform Linear Array}
\acro{UPA}{Uniform Planar Square Array}

\end{acronym}